\pdfoutput=1
\documentclass[11pt, dvipsnames]{article}

\usepackage[]{acl}

\usepackage{times}
\usepackage{latexsym}
\usepackage{float}
\usepackage{graphicx}
\usepackage{xparse}
\usepackage{xcolor}
\usepackage{soul}
\usepackage{colortbl}
\usepackage{color}
\usepackage{caption}   
\usepackage{subcaption}
\usepackage{tabularx}
\usepackage{dingbat}
\usepackage{pifont}
\usepackage{amsmath}
\usepackage{amssymb}
\usepackage{makecell}
\usepackage{arydshln}
\usepackage{stfloats}
\usepackage[T1]{fontenc}
\usepackage{titlesec}
\usepackage{enumitem}
\usepackage[utf8]{inputenc}

\usepackage{CJKutf8}
\usepackage{multirow}
\usepackage{placeins}
\usepackage{array}
\usepackage[utf8]{inputenc}
\usepackage{microtype}
\usepackage{booktabs}
\usepackage{xcolor}
\usepackage{fontawesome5}
\usepackage{caption}
\usepackage{amsfonts}
\usepackage{inconsolata}
\usepackage{arydshln}
\usepackage{CJKutf8}
\usepackage{fancyvrb}
\usepackage{listings}
\usepackage{hyperref}
\usepackage[most]{tcolorbox}

\newcommand{\llada}{\textcolor{green!50!black}{Llada-8B}}
\newcommand{\dream}{\textcolor{green!50!black}{Dream-7B}}
\newcommand{\fdllm}{\textcolor{green!50!black}{FdLLM-7B}}
\newcommand{\dvar}{\textcolor{green!50!black}{DVar-8B}}
\newcommand{\dllms}{\textcolor{green!50!black}{dLLMs}}

\newcommand{\qwen}{\textcolor{red!75!black}{Qwen-8B}}
\newcommand{\ministral}{\textcolor{red!75!black}{Ministral-8B}}
\newcommand{\llms}{\textcolor{red!75!black}{LLMs}}

\newcommand{\TAOrow}[2]{%
\textbf{#1:} & #2 \\%
}

\definecolor{myblue}{rgb}{0.82, 0.94, 0.75}

\title{The Bitter Lesson of Diffusion Language Models for Agentic Workflows: \\A Comprehensive Reality Check}

\usepackage{pifont}

\usepackage{pgfplots}
\pgfplotsset{compat=1.17}
\usetikzlibrary{shapes.geometric, arrows, positioning, shadows}
\usetikzlibrary{shapes, arrows.meta, positioning, calc}

\author{
Qingyu~Lu$^{\diamondsuit\text{\ding{95}}\dagger}$,
Liang Ding$^{\heartsuit\dagger}$,
Kanjian Zhang$^{\diamondsuit\spadesuit}$\thanks{~Corresponding Author.}, 
Jinxia Zhang$^{\diamondsuit}$,
Dacheng Tao$^{\text{\ding{95}}}$ \\
{\fontsize{10pt}{12pt}\selectfont $^{\diamondsuit}$Southeast University \
$^{\heartsuit}$Alibaba \
$^{\spadesuit}$Southeast University Shenzhen Research Institute} \\
{\fontsize{10pt}{12pt}\selectfont $^{\text{\ding{95}}}$College of Computing and Data Science at Nanyang Technological University, Singapore 639798} \\
\raisebox{-0.6ex}{\includegraphics[scale=0.032]{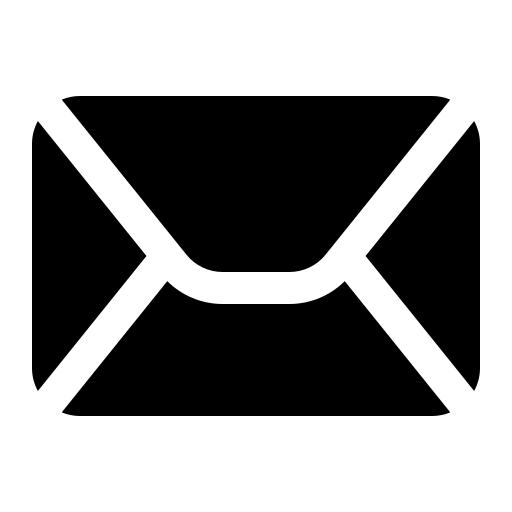}} \fontsize{9pt}{9pt}\selectfont \texttt{\{luqingyu,kjzhang,jinxiazhang\}@seu.edu.cn}, \texttt{\{liangding.liam,dacheng.tao\}@gmail.com}\\
\raisebox{-0.35ex}{\includegraphics[scale=0.03]{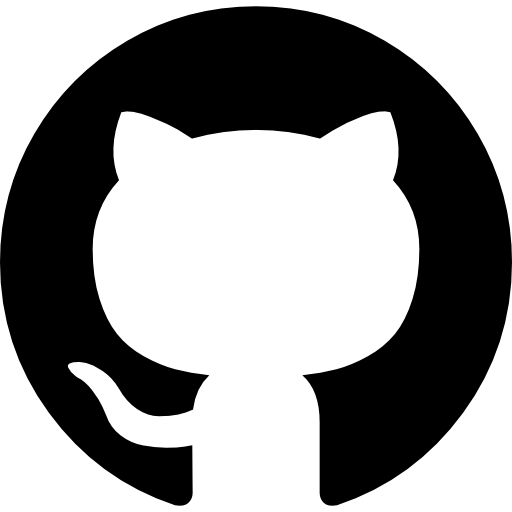}} \fontsize{9pt}{10pt}\selectfont \url{https://coldmist-lu.github.io/DiffuAgent/}}

\begin{document}
\maketitle

\begingroup
\renewcommand\thefootnote{$\dagger$}
\footnotetext{~Equal contribution. Work done while Qingyu was a visiting researcher at Nanyang Technological University.}
\endgroup

\begin{abstract}

The pursuit of real-time agentic interaction has driven interest in Diffusion-based Large Language Models (dLLMs) as alternatives to auto-regressive backbones, promising to break the sequential latency bottleneck. \textbf{However, does such efficiency gains translate into effective agentic behavior?} In this work, we present a comprehensive evaluation of dLLMs (e.g., LLaDA, Dream) across two distinct agentic paradigms: \textit{Embodied Agents} (requiring long-horizon planning) and \textit{Tool-Calling Agents} (requiring precise formatting).

Contrary to the efficiency hype, our results on Agentboard and BFCL reveal a "\textbf{\emph{bitter lesson}}": current dLLMs fail to serve as reliable agentic backbones, frequently leading to systematic failure. \textbf{\emph{(1) In Embodied settings}}, dLLMs suffer repeated attempts, failing to branch under temporal feedback. \textbf{\emph{(2) In Tool-Calling settings}}, dLLMs fail to maintain symbolic precision (e.g. strict JSON schemas) under diffusion noise. To assess the potential of dLLMs in agentic workflows, we introduce \textbf{DiffuAgent}, a multi-agent evaluation framework that integrates dLLMs as plug-and-play cognitive cores. Our analysis shows that dLLMs are effective in non-causal roles (e.g., memory summarization and tool selection) but require the incorporation of causal, precise, and logically grounded reasoning mechanisms into the denoising process to be viable for agentic tasks.

\end{abstract}
\section{Introduction}

\begin{figure}[ht]
\centering
\includegraphics[scale=0.33]{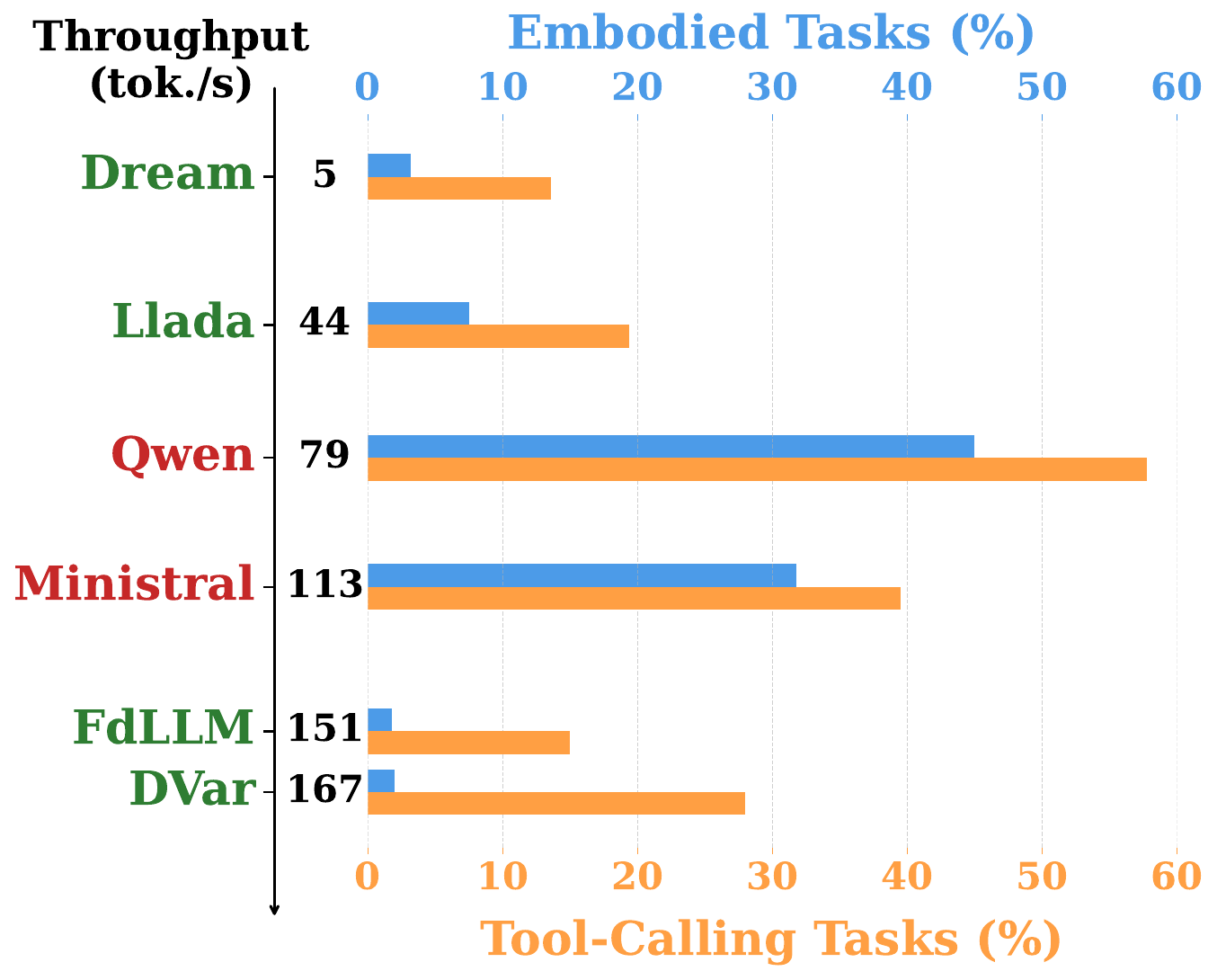}
\caption{
\textbf{Performance–Efficiency Trade-offs in Embodied and Tool-Calling Tasks.} Despite higher inference efficiency, \fdllm \ and \dvar \ do not guarantee comparable agentic performance to autoregressive \llms. \llada \ and \dream \ fall behind \llms \ in both task performance and efficiency.}
\label{fig:throughput}
\end{figure}

Agents powered by large language models (LLMs, \citealp{yang2025qwen3, jiang2024mixtral}) have demonstrated strong capabilities in planning and complex reasoning \citep{wang2024survey, luo2025large}, particularly in embodied task-solving environments \citep{feng2025multi, chang2024agentboard} and tool-calling scenarios \citep{liutoolace, patil2025bfcl}.
However, such agentic systems often suffer from a \textbf{sequential latency bottleneck}, where multi-turn interactions incur substantial inference overhead, demanding faster reasoning and more efficient decision-making.

In this context, Diffusion-based Large Language Models (dLLMs, \citealp{nie2025large}) have attracted attention as alternatives to auto-regressive backbones, owing to their higher inference efficiency enabled by parallel decoding, while maintaining competitive general performance \citep{wu2025fast2, ye2025dream}. However, \textbf{does such efficiency gains translate into effective agentic behavior?} In this work, we present a comprehensive reality check of dLLMs, focusing on their long-horizon planning capabilities as \textit{Embodied Agents} and precise formatting capabilities as \textit{Tool-Calling Agents}.

\begin{figure*}[ht]
\centering
\includegraphics[scale=0.5]{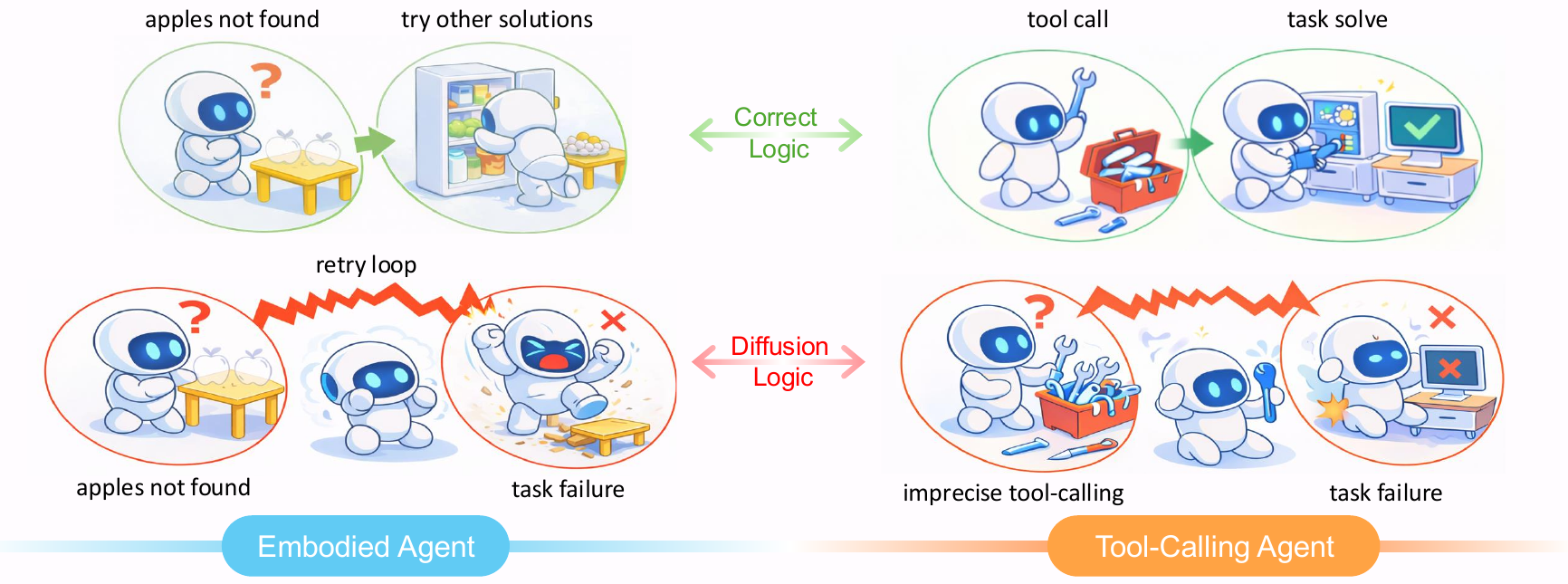}
\caption{\textbf{An Overview of Systematic Failures} in dLLMs. In embodied agent settings, dLLMs tend to repetitively retry the same action instead of exploring alternative plans. In tool-calling agent settings, imprecise or unstable tool invocation further leads to execution failures.}
\label{fig:description}
\end{figure*}

Contrary to the efficiency hype, our results (as shown in Figure~\ref{fig:throughput}) on four representative dLLMs across AgentBoard \citep{chang2024agentboard} and BFCL \citep{patil2025bfcl} reveal a \textbf{\textit{bitter lesson}}: current dLLMs fail to serve as reliable agentic backbones, particularly in multi-turn interaction scenarios, exhibiting systematic failure behaviors. Specifically, (1) \textbf{\textit{in embodied settings}}, dLLMs tend to become trapped in repetitive action loops, failing to branch into alternative plans; (2) \textbf{\textit{in tool-calling settings}}, dLLMs struggle to maintain symbolic precision when generating tool invocations, frequently violating strict JSON schemas or hallucinating API parameters, potentially due to diffusion-induced noise, as illustrated in Figure~\ref{fig:description}.

To provide deeper insights into the agentic behavior of dLLMs, we further introduce \textbf{DiffuAgent}, a novel evaluation framework that treats dLLMs as \textit{plug-and-play cognitive modules} for augmenting LLM-based agents, enabling systematic assessment under different agentic roles.
Our results show that dLLMs can be effective when deployed in non-causal roles, such as memory summarization \citep{xu2025mem, wang2025mem}, redundant trajectory detection \citep{lu-etal-2025-runaway}, and relevant tool selection \citep{liu2025toolplanner}.

Our contributions are three-fold:

\begin{itemize}

    \item We present the first systematic study of dLLMs as agentic backbones, revealing consistent and previously underexplored failure modes in multi-turn agentic reasoning.
    
    \item We propose \textbf{DiffuAgent}, the first evaluation framework that integrates dLLMs as four distinct cognitive modules within a multi-agent setting to better assess agentic behavior.

    \item We provide extensive empirical evidence showing that dLLMs are effective mainly in non-causal roles, but remain weak in causal planning and formatting–critical scenarios.

\end{itemize}

This study serves as a foundational step toward \textbf{Diffusion-native Agents}. By bridging the gap between non-autoregressive generation and agentic workflows, we highlight a promising direction for future dLLM development, enabling real-time interaction without compromising causal, precise, and logically grounded reasoning capabilities.
\section{Preliminaries}

\subsection{Embodied Agents}

Embodied agents operate in interactive environments (e.g., household or virtual worlds), where an LLM acts as the central controller, selecting actions based on accumulated interaction history. This multi-turn decision process can be formalized as a Partially Observable Markov Decision Process (POMDP), in which the agent follows a policy $\pi_\theta$ at each time step $t$ to choose the next action $a_t$:
\[
a_t \sim \pi_\theta(\cdot \mid e_{1:t-1}, u_{\rm task}),
\]
where trajectory $e_{1:t-1} = (a_1, o_1, \ldots, a_{t-1}, o_{t-1})$ consists of past actions ($a_1, \ldots, a_{t-1}$) and corresponding observations ($o_1, \ldots, o_{t-1}$), and $u_{\rm task}$ denotes contextual information associated with the specific task and initial environment configuration.

To elicit reasoning capabilities, we adopt ReAct \citep{yao2023react} for evaluating embodied agents, a widely-used agentic workflow that synergizes planning and decision-making in multi-turn interactions. This process can be formulated as:
\[
[q_t, a_t] = \pi_\theta(\cdot \mid e_{1:t-1}, u_{\rm task}),
\]
where the LLM agent generates an intermediate thought $q_t$ before producing action $a_t$. 

\subsection{Tool-Calling Agents}
Agentic LLMs are expected to exhibit effective tool-calling (also referred to as \emph{function-calling}) capabilities, where the agent is equipped with external tools and must decide whether, when, and how to invoke them to solve complex tasks \citep{patil2025bfcl}. This setting can be viewed as a special case of the agentic interaction paradigm, where, at each interaction turn, the agent is provided with a set of available tool descriptions $\mathcal{D} = \{\tau_1, \dots, \tau_N\}$ instead of interacting with a complex environment, and attempts to fulfill the user request $u_{\rm user}$ by generating one or more structured tool invocations as actions. This formats as
\begin{equation}
\mathcal{C}=\{(\tau_i,\alpha_i)\}_{i=1}^{K}\sim \pi_\theta(\cdot \mid u_{\rm user}, \mathcal{D}),
\end{equation}
where $\pi_\theta$ denotes the agent policy model, $\mathcal{C}$ is the set of generated tool calls, $\tau_i$ denotes the $i$-th selected tool, and $\alpha_i$ denotes argument. Each generated tool call is then executed by its associated tool, yielding a set of execution results:
\begin{equation}
\mathcal{O}
=
\mathrm{Exec}(\mathcal{C})
=
\{\,\tau_i(\alpha_i)\mid (\tau_i,\alpha_i)\in\mathcal{C}\,\},
\end{equation}
where the execution results $\mathcal{O}$ for all results 
are returned to the agent as feedback, based on which the agent decides whether further tool calls are required. The interaction terminates when the agent determines that the user request has been resolved or when a predefined step limit is reached.

\subsection{Diffusion-based LLMs}

Autoregressive LLMs follow a next-token prediction paradigm \citep{luo2025large}, in which tokens are decoded sequentially, one at a time. This inherent sequential nature limits decoding efficiency, particularly in generation-intensive scenarios. Inspired by diffusion probabilistic modeling \citep{yang2023diffusion}, dLLMs originally developed for continuous domains such as images \citep{amit2021segdiff} and audio \citep{nam2025diffusion}. Rather than relying on strict left-to-right generation, dLLMs generate tokens in parallel, offering greater potential for efficient inference acceleration \citep{wu2025fast}. As the parallel generation and sampling strategies differ slightly among the selected dLLMs, we explore different dLLMs and several optimization techniques during decoding stage.

\par\noindent\textit{\ding{43} See Appendix~\ref{appendix:dllm_intro} for a detailed summary of the decoding strategies of the selected dLLMs.}

\section{Experimental Setup}

\subsection{Evaluation Data}

\paragraph{Datasets}
We evaluate embodied agents using AgentBoard \citep{chang2024agentboard} across three interactive environments: AlfWorld \citep{shridharalfworld} (134 household tasks), ScienceWorld \citep{wang-etal-2022-scienceworld} (90 scientific experiments), and BabyAI \citep{chevalierbabyai} (112 grid-based navigation and interaction tasks). Tool-calling agentic ability is assessed on BFCL-v3 \citep{patil2025bfcl}. We sample at most 50 instances per BFCL-v3 category (using all samples when fewer than 50 are available), yielding 758 evaluation examples in total covering all categories.

\par\noindent\textit{\ding{43} See Appendix~\ref{appendix:benchmark} for detailed dataset descriptions, including their settings and example instances.}

\paragraph{Metrics}
For embodied agents, we report both success rate and progress rate. 
\textbf{Success rate} measures the proportion of tasks successfully completed by an agent, while \textbf{progress rate} \citep{chang2024agentboard} quantifies how much an agent advances toward the task goal, making it a more informative metric for evaluating incremental improvements. For tool-calling evaluation, we adopt the official BFCL evaluation suite and report the percentage of successful instances as our primary metric.

\subsection{Agentic Backbones}

\paragraph{\llms}
We consider open-source LLMs under 10B parameters for reproducibility and efficiency. Specifically, we use \qwen \ \citep{yang2025qwen3}, adopting the non-thinking variant to meet real-time latency constraints\footnote{\url{https://huggingface.co/Qwen/Qwen3-8B}}, and \ministral \ \citep{jiang2024mixtral}, an instruction-tuned 8B model. Both models are evaluated in text-only settings\footnote{\url{https://huggingface.co/mistralai/Ministral-3-8B-Instruct-2512}}.

\paragraph{\dllms}
We employ four recent dLLMs in our experiments: \llada \ \citep{nie2025large}, a strong diffusion LLM with general performance competitive with Llama3-8B; \dream \ \citep{ye2025dream}, which is initialized from Qwen2.5-7B weights and adopts token-level noise rescheduling for context-adaptive denoising; \fdllm \ (Fast-dLLM v2; \citealp{wu2025fast2}), a block-diffusion model enabling parallel decoding within each block for efficient inference; and \dvar \ (dLLM-Var; \citealp{yang2025diffusion}), which supports native variable-length generation via accurate EOS prediction.

\subsection{Deployment Details}
For fair comparison, we allocate one NVIDIA A800 (80GB) GPU per model. When two models are used in multi-agent scenarios (Section~\ref{sec:analysis}), two GPUs are deployed accordingly. We do not employ distributed inference. AR models (\qwen, \ministral) are deployed with vLLM \citep{kwon2023efficient} and accessed via OpenAI Chat APIs \citep{achiam2023gpt}. Diffusion LLMs (\dream, \llada, \fdllm) are reproduced using NVIDIA Fast-dLLM\footnote{\url{https://github.com/NVlabs/Fast-dLLM}} and served through FastAPI.

\subsection{Prompts}

For embodied agents, we adopt a ReAct-style prompt format for multi-turn planning and action generation. For BFCL, we follow the official implementation and build an OpenAI API version to match the input templates of different models.

\par\noindent\textit{\ding{43} See Appendix~\ref{appendix:prompts} for the prompts used.}

\begin{table*}[t]
\centering
\small
\setlength{\tabcolsep}{7pt}
\renewcommand{\arraystretch}{1.1}
\begin{tabular}{ccccccccc}
\toprule
\multirow{2}{*}[-0.7ex]{\makecell{\textbf{Embodied}\\[0.6ex]\textbf{Agent}}}
& \multicolumn{2}{c}{\textbf{AlfWorld}}
& \multicolumn{2}{c}{\textbf{ScienceWorld}}
& \multicolumn{2}{c}{\textbf{BabyAI}}
& \multicolumn{2}{c}{\textbf{Avg.}} \\ 
 \cmidrule(lr){2-3} \cmidrule(lr){4-5} \cmidrule(lr){6-7} \cmidrule(lr){8-9}
& \textbf{Success} & \textbf{Progress}
& \textbf{Success} & \textbf{Progress}
& \textbf{Success} & \textbf{Progress}
& \textbf{Success} & \textbf{Progress} \\
\midrule
\small{\qwen} 
& \textbf{76.1} {\scriptsize$\pm$0.8} & \textbf{85.6} {\scriptsize$\pm$0.3} 
& \textbf{26.7} {\scriptsize$\pm$2.2} & \textbf{55.1} {\scriptsize$\pm$0.3} 
& 32.1 {\scriptsize$\pm$0.9} & 45.7 {\scriptsize$\pm$1.0} 
& \textbf{45.0} {\scriptsize$\pm$0.2} & \textbf{62.1} {\scriptsize$\pm$0.5} \\
\small{\ministral} 
& 45.5 {\scriptsize$\pm$1.1} & 66.2 {\scriptsize$\pm$0.6} 
& 13.3 {\scriptsize$\pm$1.7} & 52.0 {\scriptsize$\pm$0.1} 
& \textbf{36.6} {\scriptsize$\pm$1.4} & \textbf{46.6} {\scriptsize$\pm$1.0} 
& 31.8 {\scriptsize$\pm$0.7} & 54.9 {\scriptsize$\pm$0.5} \\
\midrule
\small{\llada} 
& 5.2 {\scriptsize$\pm$0.4} & 18.5 {\scriptsize$\pm$0.6} 
& 1.1 {\scriptsize$\pm$0.0} & 8.6 {\scriptsize$\pm$0.4} 
& 16.1 {\scriptsize$\pm$0.9} & 22.0 {\scriptsize$\pm$1.3} 
& 7.5 {\scriptsize$\pm$0.2} & 16.4 {\scriptsize$\pm$0.4} \\
\small{\dream} 
& 0.7 {\scriptsize$\pm$0.0} & 6.0 {\scriptsize$\pm$0.1} 
& 0.6 {\scriptsize$\pm$0.6} & 5.3 {\scriptsize$\pm$0.2} 
& 8.9 {\scriptsize$\pm$0.5} & 14.8 {\scriptsize$\pm$0.5} 
& 3.4 {\scriptsize$\pm$0.1} & 8.7 {\scriptsize$\pm$0.2} \\
\small{\fdllm} 
& 3.3 {\scriptsize$\pm$3.3} & 7.8 {\scriptsize$\pm$3.3} 
& 0.7 {\scriptsize$\pm$0.6} & 6.4 {\scriptsize$\pm$1.2} 
& 5.4 {\scriptsize$\pm$0.5} & 12.6 {\scriptsize$\pm$2.2} 
& 3.1 {\scriptsize$\pm$1.3} & 8.9 {\scriptsize$\pm$1.5} \\
\small{\dvar} 
& 0.7 {\scriptsize$\pm$0.0} & 10.0 {\scriptsize$\pm$0.0} 
& 0.0 {\scriptsize$\pm$0.0} & 1.9 {\scriptsize$\pm$0.0} 
& 5.4 {\scriptsize$\pm$0.0} & 15.0 {\scriptsize$\pm$0.0} 
& 2.0 {\scriptsize$\pm$0.0} & 8.9 {\scriptsize$\pm$0.0} \\
\bottomrule
\end{tabular}
\caption{\textbf{Comparison of Success Rate (\%) and Progress Rate (\%)} across different \llms \ and 
\dllms \ on three \textit{Embodied} tasks. Best results are highlighted in \textbf{bold}. The lower-right fluctuation values indicate the variability estimated from 3 runs, showing the possible effect of error propagation.}
\label{tab:main_embodied}
\end{table*}

\begin{table*}[t]
\centering
\small
\setlength{\tabcolsep}{4.5pt}
\renewcommand{\arraystretch}{1.1}
\begin{tabular}{c c c c c c c c c c c c c}
\toprule

\multirow{2}{*}[-0.7ex]{\makecell{\textbf{Tool-Calling}\\[0.6ex]\textbf{Agent}}} &
\multicolumn{1}{c}{\textbf{Non-Live}} &
\multicolumn{5}{c}{\textbf{Single-Turn Live}} &
\multicolumn{3}{c}{\textbf{Multi-Turn}} &
\multicolumn{2}{c}{\textbf{Hallucination}} & \multirow{2}{*}[-0.5ex]{\textbf{Overall}}\\

\cmidrule(lr){2-2}
\cmidrule(lr){3-7}
\cmidrule(lr){8-10}
\cmidrule(lr){11-12}

& 
\textbf{Avg.} &
\textbf{S.} &
\textbf{M.} &
\textbf{P.} &
\textbf{PM.} &
\textbf{Avg.} &
\textbf{Standard} &
\textbf{Challenge} &
\textbf{Avg.} &
\textbf{Rel.} &
\textbf{Irrel.} & 

\\

\midrule
\qwen
& \textbf{87.5}
& \textbf{82.0} & \textbf{80.0} & \textbf{75.0} & \textbf{75.0} & \textbf{78.0}
& \textbf{20.0} & \textbf{10.0} & \textbf{12.5}
& \textbf{94.4} & 68.0
& \textbf{57.8} \\

\ministral
& 49.8
& 74.0 & 70.0 & 50.0 & 45.8 & 60.0
& 2.0 & 4.7 & 4.0
& 66.7 & 58.0
& 39.5 \\\midrule

\llada
& 23.0
& 8.0 & 26.0 & 0.0 & 12.5 & 11.6
& 0.0 & 0.0 & 0.0
& 66.7 & 56.0
& 19.4 \\

\dream
& 4.2
& 2.0 & 4.0 & 0.0 & 0.0 & 1.5
& 0.0 & 0.0 & 0.0
& 27.8 & 77.0
& 13.6 \\

\fdllm
& 1.2
& 0.0 & 0.0 & 0.0 & 0.0 & 0.0
& 0.0 & 0.0 & 0.0
& 5.6 & \textbf{99.0}
& 15.0 \\

\dvar
& 35.0
& 56.0 & 22.0 & 37.5 & 20.8 & 34.1
& 0.0 & 0.0 & 0.0
& 44.4 & 63.0
& 28.0 \\

\bottomrule
\end{tabular}

\caption{\textbf{Comparison of Success Rates (\%)} across different \llms \ and \dllms \ 
as \textit{Tool-Calling} agents.
\textbf{S.}, \textbf{M.}, \textbf{P.} and \textbf{PM.} denote simple, multiple, parallel, and a combination of parallel and multiple tool-calling tasks, respectively. Hallucination indicates whether a tool call is required (\textbf{Rel.}) or not required (\textbf{Irrel.}). Best results are in \textbf{bold}.}
\label{tab:main_toolcall}
\end{table*}

\section{Failure of \dllms \ as Agent Backbone} \label{sec:failure}

\subsection{Failure of Replan: Embodied Agents}

\paragraph{\dllms \ significantly underperform \llms}
Table~\ref{tab:main_embodied} summarizes the performance of embodied agents with LLM and dLLM backbones. 
Across all environments, dLLMs consistently underperform LLMs, achieving success rates below 10\% in most settings, with the only exception being \llada \ on BabyAI; in some cases, they fail to solve any tasks (0.0\%) in ScienceWorld. Progress-rate performance exhibits a similar pattern: almost all progress rates fall below 20\%, suggesting that dLLM agents are unable to complete even one subgoal on average.
This gap is striking given the competitive performance of dLLMs on general language benchmarks, demonstrating that such gains fundamentally fail to transfer to agentic scenarios requiring long-horizon planning.

\begin{figure}[t]
\centering
\includegraphics[scale=0.38]{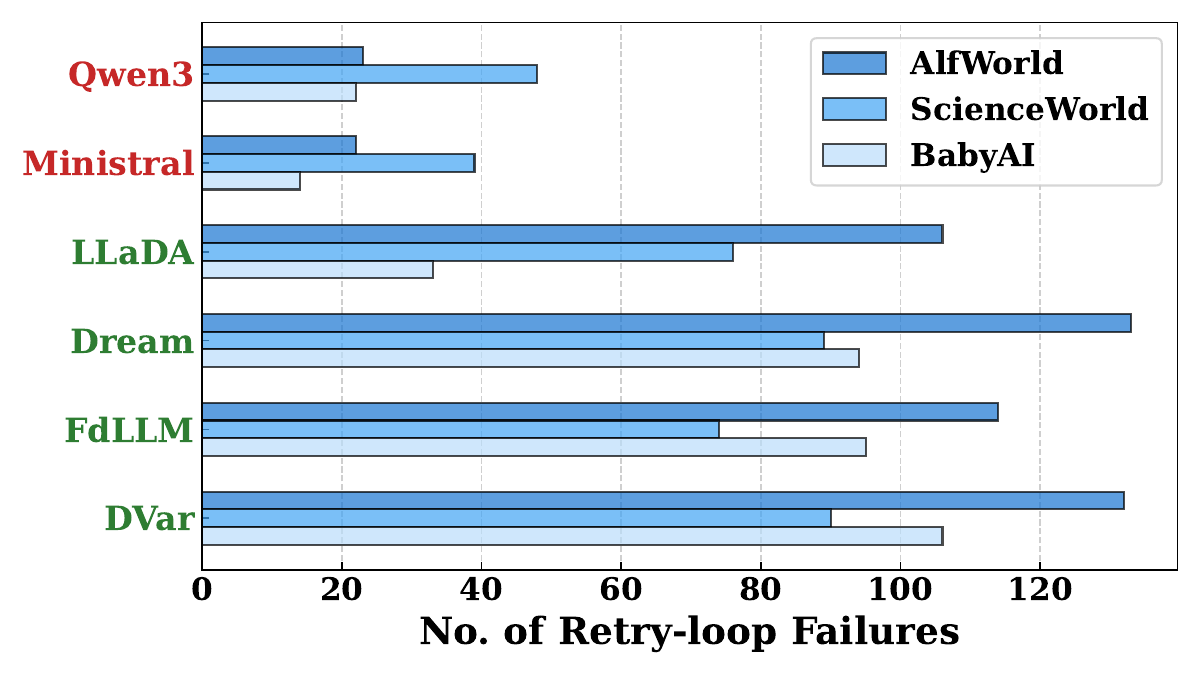}
\caption{
\textbf{Comparison of Retry-Loop Failures} across \llms \ and \dllms. A retry-loop is defined as repeatedly executing the same action for more than 3 consecutive steps during task completion.
}
\label{fig:retry}
\end{figure}

\paragraph{Retry loops as a systematic failure mode}
To further investivate the failure mode of dLLMs as embodied agents, we follow \citet{shinn2023reflexion} to define \textit{retry loop} as three or more consecutive repetitive actions and report their frequency across different backbones. As shown in Figure~\ref{fig:retry}, dLLMs exhibit significantly more frequent \emph{retry loops} than auto-regressive LLMs, repeatedly generating the same action without exploring alternatives. This indicates an over-reliance on recent context, whereas LLM-based agents exhibit more causal decision patterns and experiences \citep{sun2025seagent} by leveraging prior interactions to branch into new actions.

\paragraph{Effect of Error Propagation}
Since embodied agents may interact with the environment for up to 30 iterations, they are likely to accumulate error propagation over time. We therefore run each embodied setting for 3 runs and report the fluctuation values in Table~\ref{tab:main_embodied}. Most success-rate changes remain within 1\%, and most progress-rate fluctuations stay within 2\%. The most volatile case is \fdllm, whose success rate shows a fluctuation of $\pm 3.8\%$, suggesting that decoding changes may substantially affect the final outcome. This is consistent with Appendix~\ref{appendix:other_dllms_opt}: changing \fdllm \ from vanilla decoding to Deferred Commitment Decoding (DCD, \citealp{shu2026dcd}) raises its ALFWorld success rate from 3.3 to 10.4.\footnote{Since the BFCL test set has been rebuilt in the latest update and BFCL contains many single-turn cases, making the chance of error propagation is relatively low compared with embodied tasks. Therefore, we do not report multi-run fluctuation for BFCL.}

\subsection{Failure of Precision: Tool-Calling Agents}

\begin{figure}[t]
\centering
\includegraphics[scale=0.38]{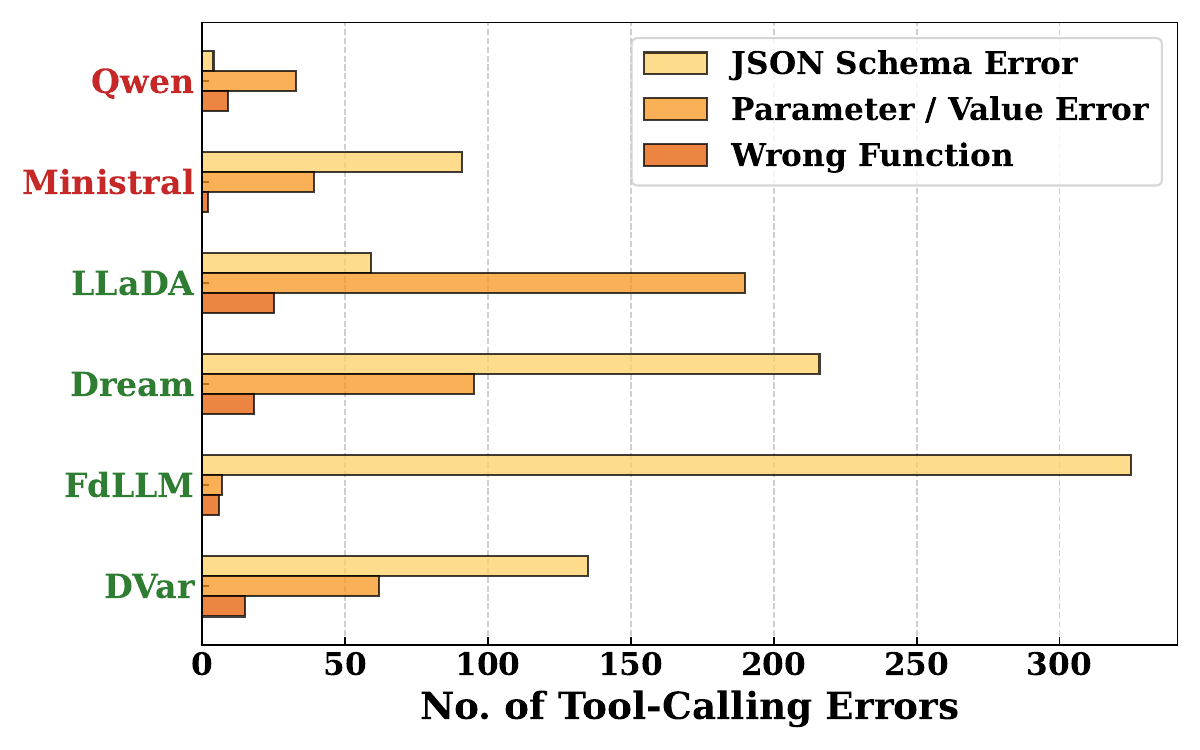}
\caption{
\textbf{Comparison of the number of tool-calling failure categories} across
\llms \ and \dllms.
}
\label{fig:tool_failure}
\end{figure}

\paragraph{\dllms \ underperform \llms \ on both single-turn and multi-turn tool-callings}
Table~\ref{tab:main_toolcall} summarizes the tool-calling results. Consistent with embodied settings, dLLMs underperform auto-regressive LLMs in both single-turn\footnote{The single-turn average is computed by excluding hallucination categories, which differs from the original BFCL implementation.} and multi-turn scenarios. Among dLLMs, \dvar \ achieves better single-turn performance but remains suboptimal. Notably, the multi-turn setting is particularly challenging for dLLMs, as none succeeds on any test instance. The high irrelevance score (Irrel.) of \fdllm \ arises from frequent incorrect tool calls that are classified as irrelevant actions.

\paragraph{Imprecise tool-call formats as a systematic failure mode}
We further summarize tool-calling failures under single-turn Live and Non-Live settings by categorizing their error types according to Abstract Syntax Tree (AST) evaluation. As shown in Figure~\ref{fig:tool_failure}, JSON schema errors and parameter/value errors dominate for both LLMs and dLLMs. However, compared to LLMs, dLLMs are more prone to produce malformed JSON schemas, except for \llada, which more often exhibits missing parameters or values. These fuzzy or ill-formed formats lead to tool execution failures, indicating that dLLMs struggle to adhere to the strict structural constraints required for tool invocation.

\subsection{Failure of Efficiency-Performance Trade-off}

As efficiency has become a key consideration for dLLMs, Figure~\ref{fig:throughput} compares efficiency and performance across models. Despite achieving high throughput (above 150 tokens/s), \fdllm \ and \dvar \ exhibit the worst embodied-task performance, with average success rates below 2\%. In contrast, auto-regressive LLMs such as \qwen \ and \ministral \ achieve stronger tool-calling and embodied reasoning performance while maintaining acceptable latency. These results show that efficiency gains in dLLMs do not directly translate into improved agentic performance.

\subsection{Bitter Lesson: Non-causal and Fuzzy Nature of \dllms}

From the above analysis, we observe a fundamental limitation of dLLMs: despite their efficiency gains, \textbf{parallel decoding weakens causal dependency and induces fuzzy intermediate states, hindering stable commitment to partial plans or structured outputs}. This aligns with established challenges in non-autoregressive generation, where the conditional independence assumption has been shown to cause `uncoordinated' structural predictions in slot filling~\cite{wu2020slotrefine} and `lexical choice errors', particularly for low-frequency tokens, in machine translation~\cite{ding2021understanding}. As a result, dLLMs perform poorly on long-horizon reasoning and strictly structured tasks, serving as a bitter lesson that they should be used with caution as backbone models in agentic workflows requiring strong temporal or symbolic consistency.

Importantly, these results do not suggest that dLLMs are ineffective in agentic systems. Since agentic tasks often require heterogeneous capabilities, we further examine the collaboration between dLLMs and LLMs in multi-agent workflows to clarify the role of dLLMs in agentic scenarios.

\par\noindent\textit{\ding{43} See Appendix~\ref{appendix:mechanistic_cases} for more rigorous definitions and theoretical explanations of the non-causal and fuzzy failure patterns of dLLMs.}

\subsection{Extended Validation}

\paragraph{dLLM Optimization Techniques}
We further examine whether recent dLLM optimization techniques could change our main conclusion. These methods improve performance, but do not close the agentic gap. In particular, Adaptive Parallel Decoding (APD) and Discrete Diffusion Forcing (D2F) bring only limited gains for \dream\ on embodied tasks, while DCD improves \fdllm\ on ALFWorld but remains limited on ScienceWorld and BabyAI. On tool-calling tasks, APD, D2F, and DCD all improve BFCL accuracy, yet a substantial gap from strong AR LLMs remains.

\paragraph{Agent-Level Optimization}
We also examine whether agent-level optimization, such as external refinement or feedback from AR LLMs, could mitigate the weakness of dLLM agents. These strategies provide limited gains and do not overturn our conclusion. AR self-refine raises the success rate from 0.7\% to 1.5\%, while periodic or step-wise AR feedback raises it to 2.2\%, and the progress rate from 10.0\% to 15.2--15.4\%. These gains suggest that external feedback can partially stabilize dLLM agents, but remain insufficient to close the gap.

\paragraph{Other Agentic Benchmarks}
We further examine whether the same limitation extends to other agentic benchmarks through Tau-Bench mock. \qwen\ is the only model that achieves a non-zero score, while the tested dLLM settings either remain at zero pass rate or fail because of response-format incompatibility.

\paragraph{Schema-Checking Methods}
We further examine whether lightweight schema-checking heuristics or structural guardrails could remove the main weakness of dLLMs in tool calling. Even when all three guardrails are combined, only 21\% of outputs achieve syntactic recovery, only 14\% reach semantic correctness, and 86\% still fail.
\par\noindent\textit{\ding{43} See Appendix~\ref{appendix:other_dllms} for detailed settings and additional experimental results.}

\section{DiffuAgent: A Multi-Agent Evaluation Framework on Analyzing Agentic Behaviors in \dllms} \label{sec:diffuagent}

\begin{figure*}[ht]
\centering
\includegraphics[scale=1.1]{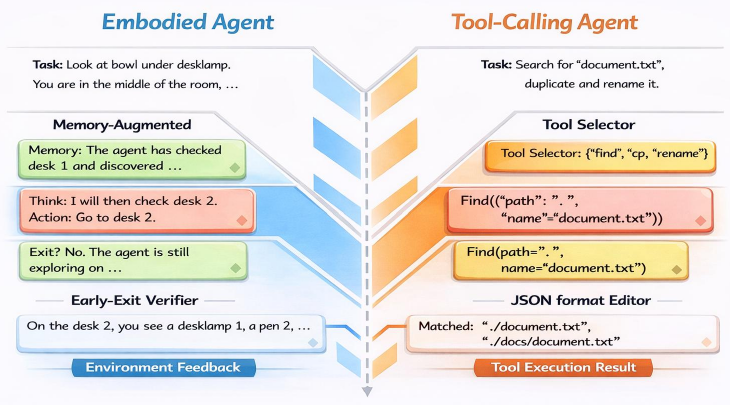}
\caption{\textbf{Overview of DiffuAgent.} The framework integrates four modules. For embodied agents, we introduce a memory-augmented module for history compression and an early-exit verifier for global trajectory checking. For tool-calling agents, we include a tool selector over the library and a JSON format editor.}
\label{fig:diffuagent}
\end{figure*}

To better understand the agentic potential of dLLMs, we introduce \textbf{DiffuAgent}, which integrates dLLMs as plug-and-play cognitive modules to augment auto-regressive LLMs. As shown in Figure~\ref{fig:diffuagent}, rather than letting dLLMs run the entire agent loop, DiffuAgent assigns them to individual functional modules, including memory, verification, and tool-related selection or format editing. This multi-agent modular design allows us to study their strengths and weaknesses more clearly, without confusing them with overall agent failures.

\subsection{Modules in Embodied Agents}

\paragraph{Pre-hoc: Memory}
We incorporate a memory-augmented module to compress long interaction histories while preserving salient information for agentic decision-making. The agent periodically summarizes past trajectories into a textual memory every $k_{\rm mem}$ steps\footnote{The memory module is invoked every \(k_{\text{mem}} = 5\) steps in our experiments, while the last two interactions are always retained to preserve recent context.}, reusing the existing memory otherwise. This process is formulated as
\begin{equation}
m' = \mathrm{Memory}\!\left(m,\, e_{t-k_{\rm mem}:t-1}, u_{\rm task}\right).
\end{equation}
During the subsequent $k_{\rm mem}$ decision steps, the policy conditions on the compressed memory together with a short-term interaction history $e_{\rm latest}$ consisting of the most recent steps:
\begin{equation}
[q_t, a_t] = \pi_\theta\!\left(\cdot \mid m',\, e_{\rm latest},\, u_{\rm task}\right).
\end{equation}
This design facilitates evaluation of agents under memory compression, where inaccurate memory updates may hinder information preservation and induce erroneous or cyclic behaviors.

\paragraph{Post-hoc: Early-Exit Verifier}

To evaluate an agent's self-awareness of being stuck, we follow \cite{lu-etal-2025-runaway} to incorporate an early-exit verification module built on an LLM or dLLM backbone. The verifier is triggered every $k_{\mathrm{earlyexit}}$ steps\footnote{We set $k_{\mathrm{earlyexit}}$ = 5 in our experiments.} and is prompted to determine whether the agent has entered a deadlock or repetitive loop. This verification process can be formulated as:
\begin{equation}
{\rm Verifier} (e_{1:t}, u_{\rm task}) \in \{0, 1\}
\end{equation}
A binary decision is then used to terminate the episode early, reducing unnecessary generation steps and improving overall efficiency.

\subsection{Modules in Tool-Calling Agents}

\paragraph{Pre-hoc: Tool Selector}
Existing tool-calling workflows suffer from a mismatch between large tool libraries and task-specific needs, which increases decision complexity and leads to inefficient or erroneous tool use. We introduce a pre-hoc tool selection module that filters the full tool set and provides a condensed subset of relevant tools prior to tool calling. Formally, at each interaction turn, given the user message $u_{\rm user}$ and the full tool set $\mathcal{D}$, the tool selector produces a reduced tool subset:
\begin{equation}
\mathcal{D'} \subseteq \mathcal{D},
\quad
\mathcal{D'} = \mathrm{Selector}(u_{\rm user}, \mathcal{D}),
\end{equation}
where $\mathcal{D'}$ contains only tools deemed relevant to the current user request. The selected tool subset $\mathcal{D'}$ is then provided to the tool-calling agent, which performs tool invocation conditioned on the reduced action space:
\begin{equation}
\mathcal{C'} \sim \pi_\theta(\cdot \mid u_{\rm user}, \mathcal{D'}).
\end{equation}

\paragraph{Post-hoc: Tool-Call Editor} Although tool calls may select correct tools and parameters, they often violate the required JSON schema, leading to execution failures. We therefore introduce a tool-call editor that post-processes malformed outputs into schema-compliant formats, enabling post-hoc evaluation of structural adherence without altering the selected function or parameters.
\section{Analysis of Agentic Behaviors in \dllms} \label{sec:analysis}

We analyze the behavior of dLLMs within the DiffuAgent framework under multi-agent settings.

\begin{table*}[t]
\centering
\small
\setlength{\tabcolsep}{5pt}
\renewcommand{\arraystretch}{1.15}
\begin{tabular}{cccccccccc}
\toprule
\multicolumn{2}{c}{\textbf{Model}}
& \multicolumn{2}{c}{\textbf{AlfWorld}}
& \multicolumn{2}{c}{\textbf{ScienceWorld}}
& \multicolumn{2}{c}{\textbf{BabyAI}}
& \multicolumn{2}{c}{\textbf{Avg.}} \\
\cmidrule(lr){1-2} \cmidrule(lr){3-4} \cmidrule(lr){5-6} \cmidrule(lr){7-8} \cmidrule(lr){9-10}
\textbf{Agent} & \textbf{Memory}
& \textbf{Success} & \textbf{Progress}
& \textbf{Success} & \textbf{Progress}
& \textbf{Success} & \textbf{Progress}
& \textbf{Success} & \textbf{Progress} \\
\midrule


\multirow{6}{*}{\qwen} & \textbf{w/o} 
& 36.6 & 59.5 & 20.0 & 53.2 & \textbf{28.6} & \textbf{39.0} & 28.4 & 50.6 \\
& \qwen 
& 54.5 & 72.8 & 28.9 & 59.5 & 21.4 & 32.2 & 34.9 & 54.8 \\
& \llada 
& \textbf{67.2} & \textbf{81.1} & \textbf{31.1} & \textbf{61.9} & 23.2 & 35.9 & \textbf{40.5} & \textbf{59.6} \\
& \dream 
& 64.9 & 77.4 & \textbf{31.1} & 60.8 & 20.5 & 33.6 & 38.9 & 57.3 \\
& \fdllm
& 57.5 & 72.8 & 28.9 & 56.5 & 20.5 & 35.2 & 35.6 & 54.8 \\
& \dvar 
& 61.2 & 76.6 & 26.7 & 58.4 & 24.1 & 35.4 & 37.3 & 56.8 \\

\cmidrule(lr){1-10}

\multirow{6}{*}{\ministral} & \textbf{w/o} 
& 22.4 & 43.7 & 17.8 & 57.6 & 33.0 & 44.2 & 24.4 & 48.5 \\
 & \ministral
& 32.8 & 60.7 & \textbf{34.4} & 65.7 & \textbf{33.9} & \textbf{45.5} & \textbf{33.7} & \textbf{57.3} \\
 & \llada
& 37.3 & 62.5 & 28.9 & \textbf{67.2} & 29.5 & 41.3 & 31.9 & 57.0 \\
 & \dream
& \textbf{39.6} & \textbf{63.5} & 26.7 & 60.9 & 25.9 & 37.3 & 30.7 & 53.9 \\
 & \fdllm
& 27.6 & 50.6 & 16.7 & 51.0 & 24.1 & 35.7 & 22.8 & 45.8 \\
 & \dvar
& 35.8 & 59.7 & 24.4 & 58.0 & 29.5 & 39.8 & 29.9 & 52.5 \\
\bottomrule
\end{tabular}
\caption{Performance comparison of \emph{memory-augmented agents} across different \llms \ and \dllms \ on three embodied environments. "\textbf{w/o}" indicates no memory, retaining only recent interactions.}
\label{tab:memory_results}
\end{table*}

\begin{figure}[t]
\centering
\includegraphics[scale=0.50]{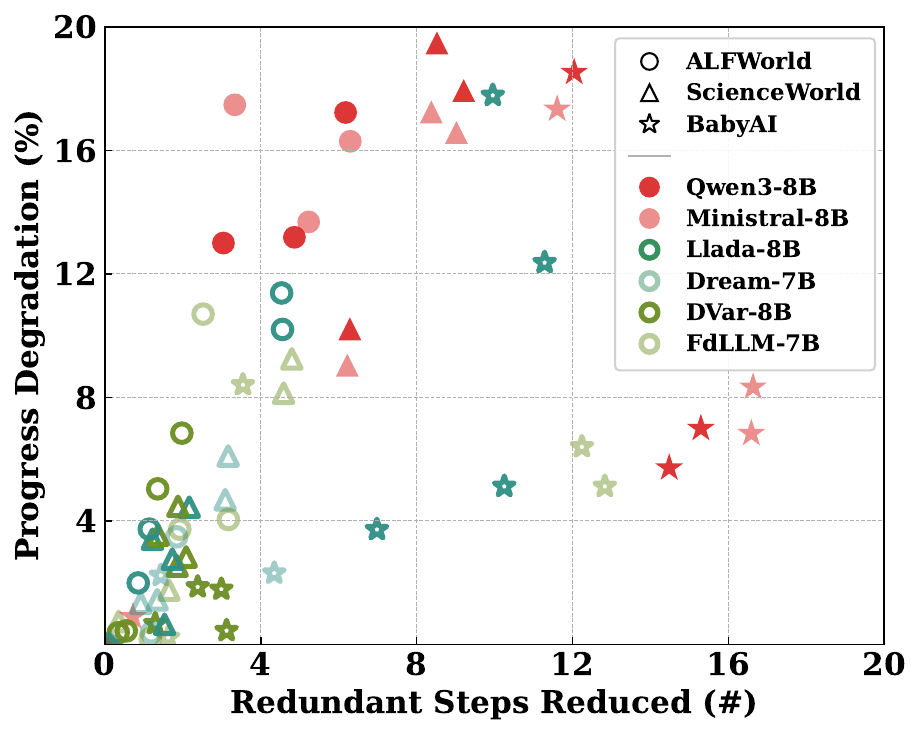}
\caption{\textbf{Comparison of early-exit behavior across LLMs and dLLMs}. Filled and hollow markers denote LLM- and dLLM-based verifiers, respectively, while different marker shapes indicate different tasks.}
\label{fig:earlyexit}
\end{figure}

\subsection{dLLMs Are Competitive Memory Modules for Memory-Augmented Agents}

In \emph{memory-augmented agents}, incorporating a memory module generally improves performance over the \textbf{w/o} baseline across tasks (Table~\ref{tab:memory_results}), indicating effective preservation of useful information. An exception is BabyAI, where long observation strings at each step may hinder effective memory summarization and lead to marginal or negative gains. Comparing memory backbones, dLLMs achieve performance comparable to \qwen, suggesting their potential as memory modules. However, performance varies across environments: \ministral \ performs better on BabyAI and ScienceWorld but worse on AlfWorld, which we attribute to its tendency to generate longer thoughts in ReAct, making summarization more challenging. This suggests that dLLMs may be less suitable for lengthy and complex reasoning traces.

\subsection{LLM Verifiers Tend to Trigger Premature Early Exits, Whereas dLLMs Terminate More Reliably}

To better assess whether dLLMs can provide early termination through self-awareness based on existing trajectories, we select the four best-performing trajectories from memory-augmented embodied agents\footnote{For the first two trajectories, we use \ministral \ and \qwen \ as both the agent and memory module, and \llada \ as the memory module for the remaining settings.} and apply early-exit verifiers implemented with different LLMs or dLLMs. We compute the two efficiency metrics, redundancy reduction and progress degradation\footnote{As illustrated in \citet{lu-etal-2025-runaway}, redundant steps quantify the potential for efficiency improvement, while progress degradation measures the loss in performance. These metrics jointly capture the trade-off between efficiency and performance.} across embodied tasks with different backbones and report them in Figure~\ref{fig:earlyexit}. An interesting phenomenon is that Auto-regressive LLMs exhibit more aggressive early exits, sharply reducing redundancy but causing severe progress loss, whereas dLLM-based verifiers behave more conservatively, achieving smaller redundancy reductions with less degradation, likely due to their global trajectory awareness.

\begin{figure*}[t]
\centering
\includegraphics[scale=0.38]{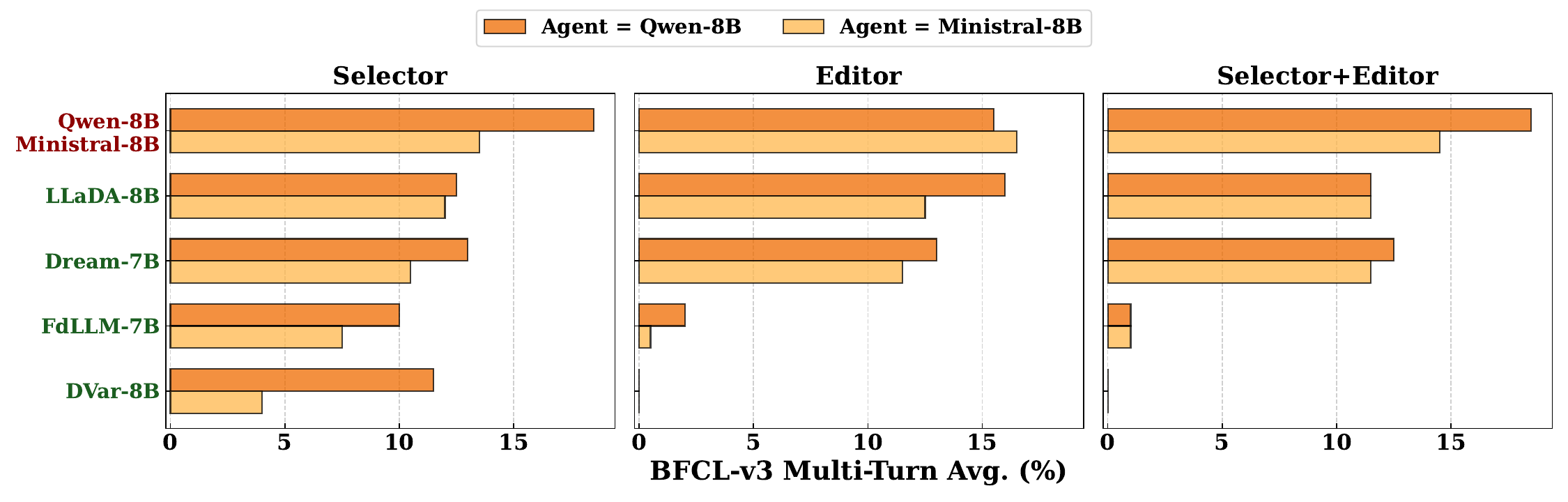}
\caption{\textbf{Ablation performance of tool-calling agents} on BFCL-v3 Multi-Turn benchmark, evaluated with different backbone models for the agent modules (\emph{Selector} and \emph{Editor}). 0 indicates no successful instance.}
\label{fig:bfcl}
\end{figure*}

\subsection{dLLMs Are Effective Tool Selectors but Struggle as Tool-Call Editors}

To further investigate the effectiveness of dLLMs as tool-calling modules, we adopt the BFCL-v3 multi-turn benchmark \citep{patil2025bfcl}, using 50 randomly selected instances to construct a 200-sample test set. The Standard setting involves multi-step tool interactions across multiple user turns, while the Challenge setting includes missing functions, missing parameters, and long-context inputs. 

We perform ablations by replacing the selector and/or editor modules in BFCL multi-turn setting. As shown in Figure~\ref{fig:bfcl}, LLM-based modules consistently outperform dLLMs. Among dLLMs, \llada \ and \dream \ serve as relatively effective selectors and editors, achieving performance comparable to LLM baselines, whereas \fdllm \ and \dvar \ degrade performance as editors, possibly due to imprecise tool calls. Interestingly, \dvar improves performance when used as a selector for \qwen \ but harms \ministral. This behavior can be attributed to \dvar’s tendency to generate weakly filtered tool subsets, benefiting \qwen \ with strong selection capacity but overwhelming \ministral \ and reducing task success.

\par\noindent\textit{\ding{43} See Appendix~\ref{appendix:bfcl_details} and Table~\ref{tab:toolcall_multi} for detailed BFCL multi-turn ablation results.}
\section{Related Work}

\paragraph{Diffusion-based LLMs}
dLLMs enable non-autoregressive generation via parallel denoising, offering substantial speedups over autoregressive LLMs. Models such as LLaDA~\citep{nie2025large} and Dream~\citep{ye2025dream} achieve competitive standalone performance, with further improvements from block-based diffusion, KV-cache reuse, and confidence-aware decoding~\citep{arriola2025interpolating,wu2025fast,wu2025fast2}. Recent decoding advances further improve diffusion inference through APD~\citep{israel2025apd}, DCD~\citep{shu2026dcd}, D2F~\citep{wang2025d2f}, and Residual Context Diffusion (RCD, \citealp{hu2026rcd}). However, dLLM behavior under multi-turn, causally grounded agentic interaction remains underexplored; we therefore systematically study dLLMs as cognitive modules within agentic workflows.

\paragraph{LLM Agents}
LLM agents show strong performance in embodied reasoning, tool use, and interactive decision-making, enabled by expert trajectory training (e.g., ETO \citep{song-etal-2024-trial}, AgentFLAN \citep{chen-etal-2024-agent}), prompt-based reasoning frameworks (e.g., ReAct \citep{yao2023react}, PreAct \citep{fu-etal-2025-preact}, StateFlow \citep{wu2024stateflow}), and planning or imitation signals \citep{lin2023swiftsage}. 
Tool-calling agents further focus on tool selection \citep{Lumer2025toolselect}, planner design \citep{liu2025toolplanner}, and schema alignment \citep{lee2025don}, but largely assume autoregressive backbones and overlook inference efficiency. 
In contrast, we investigate efficiency-oriented dLLMs as agentic backbones and reveal systematic failures in causally dependent decision processes.

\paragraph{Agent Verification and Memory}
Recent work studies model- or agent-based verification across text \citep{zheng2023judging, lu-etal-2024-error}, code \citep{chen-etal-2024-agent}, and autonomous agents \citep{panautonomous}, while memory compression methods \citep{xu2025mem,wang2025recursively, wang2025mem} recover agent capabilities under limited context windows. In contrast, we propose a multi-agent evaluation framework that treats dLLMs as cognitive modules to better measure its capabilities.

\section{Conclusion}
\label{sec:conclusion}

We conduct a systematic evaluation of dLLMs in agentic settings.
Despite their inference efficiency, we first demonstrate the "bitter lesson": dLLMs are unreliable agentic backbones under multi-turn interaction, exhibiting repetitive action loops in embodied tasks and imprecise tool calls under strict formatting constraints. We then introduce \textbf{DiffuAgent}, a modular evaluation framework that decomposes agentic workflows into plug-and-play cognitive roles. 
Our analysis shows that dLLMs struggle with causal planning and formatting-critical tasks, but remain effective in non-causal roles, such as memory summarization and tool selection, motivating diffusion-native agent designs.
\section*{Limitations}

The limitations of our work are as follows:

\begin{itemize}

    \item \textbf{Limited Coverage of dLLMs and Benchmarks:} 
    Our study evaluates a representative but limited set of diffusion-based LLMs on AgentBoard and BFCL. We focus on a subset of the test suites, considering only embodied AI tasks in AgentBoard while excluding other scenarios such as web-based tasks. For BFCL, we restrict our evaluation to versions v1–v3 to capture the core challenges of tool calling. While we believe our findings are indicative of the current behavior of dLLMs, they may not fully generalize to future architectures or broader agentic settings. We leave a more comprehensive evaluation to future work.

    \item \textbf{Inference-Only Analysis:} 
    We focus on the agentic behavior of post-training dLLMs without incorporating task-specific fine-tuning or reinforcement learning. Although this setting enables a controlled comparison, targeted training objectives or architectural adaptations may alleviate some of the observed failure modes, which we leave for future work. At the same time, prior work suggests that simple fine-tuning often behaves more like behavior cloning and may not substantially improve agentic capabilities~\cite{song-etal-2024-trial}. More meaningful gains may require multi-stage agentic training, such as continued pre-training or reinforcement learning, which incurs substantially higher computational cost.

    \item \textbf{Ablation Completeness:}
    Our ablation study covers only a subset of selector–editor configurations under the DiffAgent framework, and assumes an LLM-based main workflow agent. We focus on evaluating dLLMs as auxiliary cognitive modules rather than as primary agents. Exploring dLLMs as the main workflow agent is left for future work.
    
    \item \textbf{Fixed Agentic Workflow Assumptions:} 
    DiffuAgent evaluates dLLMs by inserting them into predefined cognitive roles within a fixed agent pipeline. This modular design facilitates systematic analysis but may underestimate the potential of diffusion models in end-to-end or co-designed agentic systems optimized for diffusion-native reasoning.

    \item \textbf{LLM Self-Awareness:}
    One possible explanation for the observed improvements is the limited self-verification capability of a single LLM, where using the same model for both generation and verification may hinder timely error detection. In contrast, multi-agent settings with heterogeneous models introduce distributional diversity that can implicitly facilitate error correction and improve agentic performance. While we acknowledge that this effect may exist, we do not believe it substantially affects our main conclusions. Due to experimental budget constraints, we do not explicitly isolate this “self-awareness” effect and leave a systematic investigation for future work.

\end{itemize}

\section*{Ethics Statement}

We take ethical considerations seriously and conduct this research in accordance with established ethical standards. This work focuses on evaluating diffusion-based large language models (dLLMs) in agentic settings and introducing a modular evaluation framework for analyzing their behaviors. The proposed evaluation framework and analyses do not introduce prompts or mechanisms intended to elicit harmful, unsafe, or deceptive outputs from the models.

All datasets, environments, and models used in this study are publicly available and widely adopted in prior research. Our experiments are conducted in simulated embodied and tool-calling environments, and no human participants are involved as evaluators or subjects. The study does not collect, process, or infer any personal or sensitive information.

Our findings highlight limitations and failure modes of dLLMs in multi-turn agentic interactions, with the goal of improving transparency, reliability, and safety in future agentic system design. We report all results and conclusions accurately and objectively, without exaggerating model capabilities or risks.
\section*{Acknowledgments}

We thank the anonymous reviewers and the area chair for their insightful comments and suggestions. This work was completed while Qingyu Lu was a visiting scholar at Nanyang Technological University, Singapore, and we thank the China Scholarship Council for its sponsorship. This research is supported by the Fundamental Research Funds for the Central Universities under Grant 2242025F20002, the National Natural Science Foundation of China under Grant 61973083, and the Shenzhen Science and Technology Program under Grant JCYJ20210324121213036. Dr Tao's research is partially supported by NTU RSR and Start Up Grants.

\bibliography{arxiv_version}

\appendix

\section{Description of dLLM Inference Strategies}
\label{appendix:dllm_intro}

This section presents a concise overview of the parallel decoding strategies of the dLLMs evaluated in this work.

\subsection{\llada: Parallel Reverse Sampling with Confidence-Based Remasking}

LLaDA \citep{nie2025large} is trained from scratch as a masked diffusion language model, learning a Transformer-based mask predictor under random masking. During SFT, prompt tokens remain unmasked while response tokens are masked; \texttt{|EOS|} is treated as a normal token during training and used for truncation at inference.

At inference, LLaDA performs reverse diffusion from a fully masked sequence. Given reverse steps \(T\) and response length \(L\), the process starts from
\[
x^{(T)} = (M,\ldots,M) \in \{M\}^L.
\]
At each step \(t = T,\ldots,1\), all masked positions ($M$) are predicted in parallel:
\[
q_i(\cdot) = p_\theta(x_i \mid p, x^{(t)}), \quad x^{(t)}_i = M,
\]
followed by sampling and partial remasking to obtain \(x^{(t-1)}\).

Instead of random remasking, LLaDA applies \emph{low-confidence remasking}, where tokens with the smallest confidence (e.g., lowest \(\max_v q_i(v)\)) are remasked, improving generation quality while maintaining parallelism. The final output is \(x^{(0)}\), truncated at the first \texttt{|EOS|}.

\subsection{\dream: Discrete Diffusion Inference and Parallel Denoising}

Dream 7B \citep{ye2025dream} follows a discrete diffusion-based generation paradigm, performing inference via iterative denoising over a fixed-length masked sequence. Starting from an initial state filled with \texttt{[MASK]} (or noise-corrupted) tokens, the model progressively refines the sequence over multiple diffusion steps, predicting all token positions in parallel at each step.

Formally, given a total of \(T\) diffusion steps, inference transforms an initial noisy sequence \(x^{(T)}\) into a clean output \(x^{(0)}\). At each step,
\[
x^{(t-1)} \sim p_\theta(x \mid x^{(t)},\, p), \quad t = T, \ldots, 1,
\]
where updates may be applied to all or a subset of masked positions according to a predefined schedule or confidence-based criterion. As multiple tokens can be updated simultaneously, Dream naturally supports parallel decoding as well as infilling-style generation.

This iterative denoising procedure exposes a flexible latency--quality trade-off: fewer diffusion steps yield faster inference at the cost of potential degradation in generation quality. The Dream implementation provides configurable step counts and sampling strategies to accommodate different deployment requirements.

\subsection{Fast-dLLM: Training-Free Acceleration via KV Caching and Confidence-Aware Parallel Decoding}

Fast-dLLM \citep{wu2025fast} is a \textbf{training-free} inference acceleration framework built on top of existing diffusion LLMs (e.g., LLaDA and Dream), modifying only the decoding procedure. It targets two challenges: enabling approximate KV caching under bidirectional attention and reducing quality degradation from parallel unmasking. In our experiments, it is applied to both \llada \ and \dream.

Fast-dLLM adopts \textbf{block-wise decoding} to make caching feasible: the generation is partitioned into blocks, where KV states of the fixed context (prompt and completed blocks) are cached and reused across denoising steps, and refreshed only at block boundaries. DualCache further improves reuse by caching both prefix and masked suffix blocks.

To mitigate the curse of parallel decoding, Fast-dLLM employs \textbf{confidence-aware parallel decoding}. For each masked position \(i\),
\[
c_i = \max_{v \in \mathcal{V}} p_\theta(x_i = v \mid \text{context}),
\]
and only tokens with \(c_i \ge \tau\) are unmasked (or at least the single highest-confidence token to ensure progress). A factor-based variant selects the largest \(K\) satisfying
\[
(K + 1)\bigl(1 - c_{(K)}\bigr) < \gamma.
\]
Together with (Dual) KV caching, this strategy achieves large inference speedups with minimal accuracy loss.

\subsection{\fdllm: Block-Diffusion Inference with Hierarchical Caching}

Fast-dLLM v2 \citep{wu2025fast2} adapts a pretrained autoregressive LLM (e.g., Qwen2.5-Instruct \citep{yang2025qwen3}) into a block-diffusion decoder via light fine-tuning. The core design enforces causal generation across blocks while allowing bidirectional refinement within each block, preserving global left-to-right semantics while enabling parallel token updates locally.

At inference time, blocks are generated sequentially. For block \(b\), the prefix \(x_{<b}\) is fixed, and decoding starts from an all-mask block
\[
x_b^{(0)} = (M, \ldots, M) \in \{M\}^B.
\]
Masked positions are iteratively refined using confidence-aware parallel decoding. For each masked index \(i\),
\[
c_i = \max_{v \in \mathcal{V}} p_\theta(x_i = v \mid x_{<b}, x_b),
\]
and tokens with \(c_i \ge \tau\) are unmasked (or at least the single highest-confidence token is selected to ensure progress).

Efficiency is achieved via \textbf{hierarchical caching}: a block-level cache reuses KV states for fully decoded past blocks, while a sub-block cache (DualCache-style prefix--suffix reuse) reduces recomputation during within-block refinement. This design achieves up to \(\sim 2.5\times\) speedup over standard autoregressive decoding with minimal quality degradation.

\subsection{\dvar: EOS-Led Variable-Length Block Diffusion}

dLLM-Var \citep{yang2025diffusion} enables native variable-length decoding for diffusion LLMs, removing the fixed-length constraint of vanilla dLLMs. It is obtained by lightly fine-tuning \llada \ to adjust EOS behavior while preserving the diffusion modeling.

At inference, decoding follows an EOS-led block-diffusion process under bidirectional attention. Starting from a prompt, a block of \(B\) masked tokens is appended and denoised in parallel; if no \texttt{|EOS|} is produced, additional masked blocks are appended iteratively until EOS appears. At block \(k\),
\[
x^{(k)} = [x_{\text{prompt}}, \hat{x}_{1:(k-1)B}, \underbrace{M, \ldots, M}_{B}],
\]
and decoding terminates at the earliest EOS position.

Efficiency comes from simple KV caching: the prompt and completed blocks are cached as fixed context, while computation focuses on the current block. Parallel updates can be gated by a confidence threshold,
\[
\max_v p_\theta(x_i = v) \ge \tau \ (\tau \approx 0.9),
\]
achieving substantial speedups without specialized attention masks or complex cache refresh.

\section{Description of Benchmarks} \label{appendix:benchmark}

We provide a brief description of the benchmarks used in our experiments to facilitate a clear understanding of the task requirements and to improve reproducibility and comparability with prior work.

\subsection{AgentBoard}

We adopt three embodied AI task environments from AgentBoard \citep{chang2024agentboard}: AlfWorld, ScienceWorld, and BabyAI. Table~\ref{tab:example_agentboard} reports the number of tasks ("\#Tasks") in each test set, along with a representative example trajectory from each environment, collected using \qwen\ as the agentic backbone.

\subsection{BFCL-v3}

We evaluate tool-calling agents on the widely used tool-calling benchmark BFCL-v3 \citep{patil2025bfcl}. Table~\ref{tab:example_bfcl} summarizes the detailed configuration of the test set. Since the original BFCL test set contains over 4,000 samples, we randomly select 50 samples from each category defined in the original dataset (see the “Categories II” column) using a fixed random seed of 42. For categories with fewer than 50 samples, we include all available samples for evaluation. This procedure yields a total of 758 test samples. To accommodate space constraints, we further regroup the original categories into a set of consolidated labels (“Categories I”) when reporting the main results in Table~\ref{tab:main_toolcall}.

\section{Definitions and Theoretical Explanations of the Non-Causal and Fuzzy Nature of \dllms}
\label{appendix:mechanistic_cases}

To support the analysis in Section~4.4, we provide more rigorous definitions, case demonstrations, and theoretical explanations for the non-causal and fuzzy failure patterns of dLLMs.

\subsection{Non-Causal}

\paragraph{Definition}

\textit{Non-causal} refers to the contrast with standard autoregressive (AR) causal decoding, where tokens are generated sequentially based on previous context (\(p(x_t \mid x_{<t})\)), enabling reasoning and Chain-of-Thought (CoT) capabilities. Diffusion-based LLMs (dLLMs) generate tokens in parallel, which may weaken this causal assumption and reduce the model's ability for incremental reasoning.

\paragraph{Theoretical Explanation}

\begin{table*}[t]
\centering
\small
\renewcommand{\arraystretch}{1.15}
\begin{tabularx}{0.92\textwidth}{>{\centering\arraybackslash}p{0.08\textwidth} >{\raggedright\arraybackslash}p{0.44\textwidth} >{\raggedright\arraybackslash}p{0.12\textwidth} >{\raggedright\arraybackslash}p{0.20\textwidth}}
\toprule
\textbf{Step} & \textbf{Thought} & \textbf{Action} & \textbf{Observation} \\
\midrule
1 & I should start by checking the cabinets for finding a pan. & go to cabinet 1 & On cabinet 1, you see a bowl 1 and a plate 1. \\
2 & I should start by checking the cabinets for finding a pan. & go to cabinet 1 & Nothing happens. \\
3 & I should start by checking the cabinets for finding a pan. & go to cabinet 1 & Nothing happens. \\
\bottomrule
\end{tabularx}
\caption{\textbf{Retry loop example} from FastdLLM-7B on AlfWorld \#44. \textbf{Goal:} Cool a pan and put it on the countertop.}
\label{tab:retry_case}
\end{table*}

\begin{table*}[t]
\centering
\small
\renewcommand{\arraystretch}{1.15}
\begin{tabularx}{0.92\textwidth}{>{\raggedright\arraybackslash}p{0.22\textwidth} >{\raggedright\arraybackslash}p{0.66\textwidth}}
\toprule
\textbf{Field} & \textbf{Content} \\
\midrule
\textbf{User Query} & Can you search for the song `Shape of You'? \\
\textbf{Function} & \texttt{play\_spotify\_song} --- searches for a song on Spotify using a query and plays it. \\
\textbf{Model Output (Raw)} & \texttt{[play\_spotify\_song(query',':': 'Shape of You',} \\
\textbf{Correct Function Call} & \texttt{[play\_spotify\_song(query="Shape of You")]} \\
\bottomrule
\end{tabularx}
\caption{\textbf{Schema violation example} from FastdLLM-7B on BFCL live\_simple\_239-125-2.}
\label{tab:schema_case}
\end{table*}

As shown in Table~\ref{tab:retry_case}, dLLMs may repeatedly preserve an outdated action hypothesis even after receiving new environmental feedback. In AR decoding, the causal attention mask ensures that each position only attends to previous tokens, so new observations (e.g., ``On cabinet 1, see a bowl and a plate'') directly and exclusively update all subsequent actions, allowing the agent to correct its plan when previous attempts fail. In diffusion decoding, however, full bidirectional attention is applied across the entire sequence, diluting the influence of any single observation and weakening the model's sensitivity to local feedback. As a result, the model may repeat the same action (``go to cabinet 1'') despite receiving negative environmental feedback, leading to failure modes such as retry loops in embodied agent tasks.

\subsection{Fuzzy}

\paragraph{Definition}

\textit{Fuzzy} refers to the tendency of dLLMs to produce structurally malformed outputs due to their continuous-to-discrete generation process. Unlike AR decoding, which enforces local syntactic constraints token by token, dLLMs iteratively refine continuous latent representations of the entire sequence before projecting to discrete tokens, making them prone to schema violations such as malformed function calls or broken key-value syntax.

\paragraph{Theoretical Explanation}

As shown in Table~\ref{tab:schema_case}, unlike AR decoding, where the causal attention mask ensures each structural token (e.g., \texttt{=}, \texttt{"}) is uniquely constrained by its prefix, diffusion decoding applies bidirectional attention across the entire sequence, lacking position-specific syntactic constraints. This leads to structurally inconsistent outputs --- the model produces garbled syntax (\texttt{query',':': 'Shape of You'}) instead of the valid schema (\texttt{query="Shape of You"}), rendering the function call unparseable despite correctly identifying the target function and argument.

\section{Discussion of Other dLLMs and Optimization Techniques}
\label{appendix:other_dllms}

This appendix discusses three additional questions raised during the rebuttal stage that are closely related to the scope and interpretation of our main results.

\subsection{Can dLLM Optimization Techniques Solve Failures in Agentic Workflows?}
\label{appendix:other_dllms_opt}

\paragraph{Motivation}
Our main experiments focus on a restricted set of dLLM optimization strategies (e.g., FastdLLM and DLLMVar). To broaden this coverage, we further incorporate several recent decoding advances as supplementary experiments and examine whether they materially change the conclusions of our main study.

\paragraph{Settings}
For clarity, we list the relevant methods below together with the implementation settings adopted in our evaluation:
\begin{itemize}[leftmargin=1.5em]
    \item \textbf{APD}~\citep{israel2025apd}: We implement \dream\ + APD using the balanced configuration to preserve both performance and efficiency ($R=0.7$, $W=16$, $M=100$), with Qwen2.5-0.5B as the approximate AR model.
    \item \textbf{DCD}~\citep{shu2026dcd}: We adopt DCD with \fdllm\ (dual cache), corresponding to the best-performing configuration reported in the original paper.
    \item \textbf{D2F}~\citep{wang2025d2f}: We apply D2F on top of \dream, as this combination achieves the strongest performance in the original study.
\end{itemize}

\begin{table*}[t]
\centering
\small
\setlength{\tabcolsep}{6pt}
\renewcommand{\arraystretch}{1.15}
\begin{tabular}{llccccccc}
\toprule
\multirow{2}{*}{\textbf{dLLM}} & \multirow{2}{*}{\textbf{Inference}}
& \multicolumn{2}{c}{\textbf{AlfWorld}}
& \multicolumn{2}{c}{\textbf{ScienceWorld}}
& \multicolumn{2}{c}{\textbf{BabyAI}}
& \multicolumn{1}{c}{\textbf{BFCL}} \\
\cmidrule(lr){3-4} \cmidrule(lr){5-6} \cmidrule(lr){7-8} \cmidrule(lr){9-9}
& & \textbf{Success} & \textbf{Progress} & \textbf{Success} & \textbf{Progress} & \textbf{Success} & \textbf{Progress} & \textbf{Single-Live} \\
\midrule
\multirow{3}{*}{\dream} & Vanilla & 0.7 & 6.0 & 0.6 & 5.3 & 8.9 & 14.8 & 1.5 \\
 & APD & 0.0 & 10.9 & 0.0 & 3.3 & 4.5 & 9.4 & 31.4 \\
 & D2F & 2.9 & 13.8 & 2.2 & 6.7 & 9.8 & 23.6 & 34.3 \\
\multirow{2}{*}{\fdllm} & Vanilla & 3.3 & 7.8 & 0.7 & 6.4 & 5.4 & 12.6 & 0.0 \\
 & DCD & 10.4 & 30.8 & 0.0 & 6.4 & 8.0 & 13.3 & 30.7 \\
\bottomrule
\end{tabular}
\caption{\textbf{Comparison of recent dLLM optimization techniques} on embodied and tool-calling tasks. We report Success Rate (\%) and Progress Rate (\%) on ALFWorld, ScienceWorld, and BabyAI, and Accuracy (\%) on BFCL Single-Live.}
\label{tab:other_dllm_optimizations}
\end{table*}

\paragraph{Results}
From Table~\ref{tab:other_dllm_optimizations}, we observe that on embodied tasks, APD and D2F yield only limited gains for \dream. On ALFWorld, they modestly improve progress rate while leaving success rate low, whereas on ScienceWorld and BabyAI, D2F is consistently stronger than APD. For \fdllm, DCD substantially improves ALFWorld and BFCL performance, but its gains on ScienceWorld and BabyAI remain limited. On tool-calling tasks, APD, D2F, and DCD all raise the original baseline to above 30\% accuracy, placing them in a similar performance tier to the stronger dLLM variants in our main results. Overall, although these techniques lead to notable improvements, a substantial gap between dLLMs and AR LLMs remains, and our overall conclusion still holds.

\subsection{Can Agent-Level Optimization Improve dLLM Agent Performance?}
\label{appendix:agent_level_opt}

\paragraph{Motivation}
Our main experiments do not consider two related directions. First, we do not cover agentic workflows beyond standard ReAct-style execution, such as Self-Refine~\citep{madaan2023selfrefine} and Reflexion~\citep{shinn2023reflexion}. Second, we do not study the use of AR models as feedback signals to assist a dLLM agent, but only include AR models as sub-modules within the DiffuAgent framework. A reasonable concern is that if AR feedback can help a dLLM agent converge to a reasonable multi-step action chain, then the weakness of dLLMs may reflect unstable planning rather than a systematic defect. We emphasize that incorporating techniques such as multi-trial refinement (e.g., self-refine) or in-trajectory autoregressive feedback is beyond the main scope of this paper, since it shifts the focus from evaluating native dLLM behavior to optimizing agent-level performance and also reduces the runtime efficiency that motivates the use of dLLMs. Nevertheless, examining such settings provides a more complete validation.

\paragraph{Settings}
To examine this concern, we conduct additional experiments on the ALFWorld test set. We use \dvar\ as the primary agent backbone and \qwen\ \citep{yang2025qwen3} as the AR refinement or feedback module. We consider three augmentation settings:
\begin{enumerate}[leftmargin=1.5em]
    \item AR self-refine with up to 2 trials, where the AR model performs reflection and replanning based on the trajectory generated by the dLLM;
    \item AR feedback every 5 steps within a single trajectory;
    \item AR feedback at every step, forming a tightly coupled hybrid loop.
\end{enumerate}

\begin{table}[t]
\centering
\small
\renewcommand{\arraystretch}{1.15}
\begin{tabular}{lcc}
\toprule
\textbf{Agent Optimization Method} & \textbf{SR (\%)} & \textbf{PR (\%)} \\
\midrule
\dvar\ (Baseline) & 0.7 & 10.0 \\
+ AR Self-Refine (2 trials) & 1.5 & 13.5 \\
+ AR Feedback (every 5 steps) & 2.2 & 15.2 \\
+ AR Feedback (every step) & 2.2 & 15.4 \\
\bottomrule
\end{tabular}
\caption{\textbf{Comparison of agent-level optimization methods} on ALFWorld. We report Success Rate (\%) and Progress Rate (\%) with \dvar\ as the primary dLLM backbone and \qwen\ as the AR refinement or feedback module.}
\label{tab:beyond_react_methods}
\end{table}

\paragraph{Results}
Table~\ref{tab:beyond_react_methods} shows that these agent-level optimization strategies do improve embodied performance, but the gains remain limited. AR self-refine raises the success rate from 0.7\% to 1.5\%, while periodic or step-wise AR feedback increases it to 2.2\%. The progress rate also rises from 10.0\% to 15.2--15.4\%. These results suggest that external autoregressive refinement can partially mitigate the identified failure modes, but it still does not close the gap.

\subsection{Can the Bitter Lesson Generalize to Other Agentic Benchmarks?}
\label{appendix:tau_bench_mock}

\paragraph{Motivation}
Our main experiments focus on BFCL as the primary tool-calling benchmark and therefore do not directly cover conversational multi-turn tool-use settings such as Tau-Bench~\citep{yao2024taubench}. In our preliminary attempts on Tau-Bench and SWE-Bench~\citep{jimenez2023swebench}, we found that most samples exceed the context window of the tested dLLMs, making direct evaluation invalid. Therefore, we only report Tau-Bench mock, the lightest Tau-Bench domain, as a basic sanity-check benchmark for lightweight interactive tool use, and leave broader evaluation on larger Tau-Bench domains and SWE-Bench to future work.

\paragraph{Settings}
We evaluate on the Tau-Bench mock with max\_steps=20 and max\_errors=5. Since the original split contains one broken task (update\_task\_with\_user\_tools), we exclude it and report results on the remaining 9 tasks. We compare \qwen, \fdllm\ with DCD, \dream\ with D2F, and \dvar.

\begin{table}[t]
\centering
\small
\setlength{\tabcolsep}{3pt}
\renewcommand{\arraystretch}{1.15}
\begin{tabular}{lccccc}
\toprule
\textbf{Model} & \textbf{Status} & \textbf{Eval.} & \textbf{Reward} & \textbf{Pass} & \textbf{Infra} \\
\midrule
\qwen & valid & 8 & 0.25 & 0.25 & 1 \\
\fdllm{} + DCD & valid & 9 & 0.00 & 0.00 & 0 \\
\dream{} + D2F & invalid & 0 & - & - & 9 \\
\dvar & valid & 9 & 0.00 & 0.00 & 0 \\
\bottomrule
\end{tabular}
\caption{\textbf{Results on Tau-Bench mock.} We report the number of evaluated tasks, average reward, pass rate, and infrastructure errors after excluding one broken task from the original 10-task split, leaving 9 effective tasks.}
\label{tab:tau_bench_mock}
\end{table}

\paragraph{Results}
Table~\ref{tab:tau_bench_mock} shows that \qwen\ is the only model that achieves a non-zero score on Tau-Bench mock, indicating that the benchmark pipeline and text-based tool-calling fallback are workable. In contrast, \fdllm\ with DCD and \dvar\ both finish valid runs but remain at zero pass rate, typically producing repetitive natural-language responses instead of entering the required tool-calling protocol. \dream\ with D2F fails earlier in the stack, with all 9 tasks ending as infrastructure errors because the benchmark receives assistant turns with neither usable content nor tool calls. Overall, these results suggest that the main bottleneck is not only serving stability, but also policy following and action formatting under the Tau-Bench interaction loop, which further strengthens our bitter lesson.

\subsection{Can Schema-Checking Methods Improve dLLM Tool Calling?}
\label{appendix:schema_checking}

\paragraph{Motivation}
Our main experiments do not separately evaluate whether simple schema-checking heuristics or lightweight structural guardrails could materially close the tool-calling gap. A natural concern is that if such post-hoc repairs were already sufficient to recover most failures, then the observed weakness of dLLMs might mainly come from superficial formatting errors rather than deeper decoding problems.

\paragraph{Settings}
To assess whether lightweight structural guardrails could close the tool-calling performance gap, we conduct an additional controlled study on 100 uniformly sampled BFCL outputs across four dLLMs. We consider three simple guardrails:
\begin{itemize}
    \item G1: Repetition truncation, which removes repetitive trailing content;
    \item G2: First-bracket extraction, which extracts the first complete bracketed structure from the raw output;
    \item G3: Dict-unpacking repair, which repairs malformed Python-style dict-unpacking patterns.
\end{itemize}
We also evaluate their combined version.

\begin{table}[t]
\centering
\small
\setlength{\tabcolsep}{4pt}
\renewcommand{\arraystretch}{1.15}
\begin{tabular}{lccc}
\toprule
\textbf{Guardrail} & \textbf{Parse (\%)} & \textbf{Semantic (\%)} & \textbf{Fail (\%)} \\
\midrule
G1: Truncate & 9 & 8 & 92 \\
G2: Extract & 20 & 12 & 88 \\
G3: Repair & 1 & 1 & 99 \\
G1+G2+G3 & 21 & 14 & 86 \\
\bottomrule
\end{tabular}
\caption{\textbf{Results of lightweight schema guardrails on BFCL outputs.} Each number denotes the percentage of sampled cases in the corresponding category over 100 BFCL outputs. Parse denotes syntactic recovery, Semantic denotes semantic correctness, and Fail denotes the remaining failures. G1 denotes truncation, G2 denotes first-bracket extraction, and G3 denotes dict-unpacking repair.}
\label{tab:schema_checking_methods}
\end{table}

\paragraph{Results}
Table~\ref{tab:schema_checking_methods} shows that even when combining all three guardrails (G1+G2+G3), only 21\% of BFCL outputs achieve syntactic recovery and only 14\% reach semantic correctness, while 86\% still fail. Among these remaining failures, 79\% still break at the parsing stage due to deeply corrupted or incomplete AST structures that cannot be repaired by local structural heuristics, while the remaining 7\% are syntactically valid but semantically incorrect, such as wrong function names, incorrect argument values, or mismatched argument counts. We further note that BFCL already evaluates schema adherence through its built-in \texttt{ast\_decoder} verification. This means the benchmark itself already provides a strong structural validity check, and our additional analysis shows that the dominant failure modes arise from deeper decoding and semantic errors rather than superficial format violations.

\section{Detailed Results on the BFCL-v3 Benchmark under the DiffuAgent Framework} \label{appendix:bfcl_details}


To facilitate clearer comparisons and ensure reproducibility, we report detailed statistical results on the BFCL-v3 benchmark in Table~\ref{tab:toolcall_multi}, reported separately for the “Standard” and “Challenge” categories. The results are consistent with Figure~\ref{fig:bfcl}.

\newcommand{\avgcell}[1]{\hspace{1em}#1\hspace{1em}}

\begin{table*}[t]
\centering
\small
\setlength{\tabcolsep}{2pt}
\renewcommand{\arraystretch}{1.2}
\begin{tabular}{cccccccccc}
\toprule
\multicolumn{5}{c}{\textbf{\qwen \ Agent}} & 
\multicolumn{5}{c}{\textbf{\ministral \ Agent}} \\
\cmidrule(lr){1-5} \cmidrule(lr){6-10}

\multicolumn{2}{c}{\textbf{Modules}} & \multicolumn{3}{c}{\textbf{BFCL Multi-Turn (\%)}} & 
\multicolumn{2}{c}{\textbf{Modules}} & \multicolumn{3}{c}{\textbf{BFCL Multi-Turn (\%)}} \\
\cmidrule(lr){1-2} \cmidrule(lr){3-5} \cmidrule(lr){6-7} \cmidrule(lr){8-10}

\textbf{Selector} & \textbf{Editor} & \textbf{Standard} & \textbf{Challenge} & \avgcell{\textbf{Avg.}} &
\textbf{Selector} & \textbf{Editor} & \textbf{Standard} & \textbf{Challenge} & \avgcell{\textbf{Avg.}} \\
\midrule

\multicolumn{10}{c}{\textbf{\textit{Agent + Selector}}} \\
\hdashline[3pt/3pt]
\noalign{\vskip 0.5ex}

\qwen      & - & \textbf{26.0} & \textbf{16.0} & \textbf{18.5}
           & \ministral & - & \textbf{18.0} & \textbf{12.0} & \textbf{13.5} \\

\llada     & - & 16.0 & 11.3 & 12.5
           & \llada     & - & 16.0 & 10.7 & 12.0 \\

\dream     & - & 14.0 & 12.7 & 13.0
           & \dream     & - & 12.0 & 10.0 & 10.5 \\

\fdllm     & - & 16.0 & 8.0  & 10.0
           & \fdllm     & - & 12.0 & 6.0  & 7.5  \\

\dvar      & - & 20.0 & 8.7  & 11.5
           & \dvar      & - & 2.0  & 4.7  & 4.0  \\

\midrule
\multicolumn{10}{c}{\textbf{\textit{Agent + Editor}}} \\
\hdashline[3pt/3pt]
\noalign{\vskip 0.5ex}

- & \qwen  & \textbf{26.0} & 12.0 & 15.5
  & - & \ministral & \textbf{14.0} & \textbf{17.3} & \textbf{16.5} \\

- & \llada & 20.0 & \textbf{14.7} & \textbf{16.0}
  & - & \llada     & 12.0 & 12.7 & 12.5 \\

- & \dream & 16.0 & 12.0 & 13.0
  & - & \dream     & 12.0 & 11.3 & 11.5 \\

- & \fdllm & 2.0  & 2.0  & 2.0
  & - & \fdllm     & 0.0  & 0.7  & 0.5  \\

- & \dvar  & 0.0  & 0.0  & 0.0
  & - & \dvar      & 0.0  & 0.0  & 0.0  \\

\midrule
\multicolumn{10}{c}{\textbf{\textit{Agent + Selector + Editor}}} \\
\hdashline[3pt/3pt]
\noalign{\vskip 0.5ex}

\qwen  & \qwen  & \textbf{28.0} & \textbf{15.3} & \textbf{18.5}
       & \ministral & \ministral & \textbf{20.0} & \textbf{12.7} & \textbf{14.5} \\

\llada & \llada & 16.0 & 10.0 & 11.5
       & \llada & \llada & 16.0 & 10.0 & 11.5 \\

\dream & \dream & 22.0 & 9.3  & 12.5
       & \dream & \dream & 12.0 & 11.3 & 11.5 \\

\fdllm & \fdllm & 2.0  & 0.7  & 1.0
       & \fdllm & \fdllm & 2.0  & 0.7  & 1.0 \\

\dvar  & \dvar  & 0.0  & 0.0  & 0.0
       & \dvar  & \dvar  & 0.0  & 0.0  & 0.0 \\

\bottomrule
\end{tabular}
\caption{\textbf{Ablation performance of tool-calling agents} on BFCL-v3 Multi-Turn benchmark, evaluated with different backbone models for the agent modules (\emph{Selector} and \emph{Editor}).}
\label{tab:toolcall_multi}
\end{table*}
\definecolor{EmbodiedColor}{HTML}{6495ed}
\definecolor{ToolcallColor}{HTML}{F4A261}

\section{Prompt Context}
\label{appendix:prompts}

\subsection{Embodied Agentic Workflow}

\paragraph{ReAct Prompt}
Following \citet{chang2024agentboard}, we use the provided task instruction, task goal, and in-context example for each dataset. As \citet{chang2024agentboard} adopt an act-only prompting style rather than ReAct prompting, we follow \citet{song-etal-2024-trial} to design a ReAct-style prompt format. For ALFWorld and ScienceWorld, we include valid action lists for these two datasets to ensure fair comparison with prior work \citep{song-etal-2024-trial, fuagentrefine}. See the corresponding  \hyperlink{box:embodied_react}{\texttt{PROMPT}} with AlfWorld as an example.

\paragraph{DiffuAgent: Memory} For memory-augmented agents, we use \textit{gpt-5.1-2025-11-13} to construct 3 memory \hyperlink{box:memory_exemplars}{\texttt{EXEMPLARS}} by transforming original ReAct trajectories, and using a direct \hyperlink{box:memory_update}{\texttt{PROMPT}} for summarizing the previous memory.

\paragraph{DiffuAgent: EarlyExit} We follow \citep{lu-etal-2025-runaway} to use the same early-exit \hyperlink{box:early_exit}{\texttt{PROMPT}} for external verification.

\subsection{Tool-Calling Agentic Workflow}

\paragraph{Tool-Calling Prompt} 
For the experiments in this work, we follow BFCL \citep{patil2025bfcl} and adopt their \hyperlink{box:tool_call}{\texttt{PROMPT}} without any modification to ensure faithful reproduction. Although \qwen\ provides a dedicated tool-calling mode, we do not use it in this work, as our initial experiments indicate that many tools provided in BFCL do not strictly adhere to the requirements of the tool-calling mode.

\paragraph{DiffuAgent: Tool Selector}
We employ a tool selector prior to tool-call generation; the corresponding prompt is provided in \hyperlink{box:tool_selector}{\texttt{PROMPT}}. The selector takes the available functions and the interaction history as input, and outputs a subset of tools to be selected. We then update the tool descriptions to include only this selected subset.

\paragraph{DiffuAgent: Too-Call Editor}
We introduce a tool editor to refine generated tool calls; the corresponding prompt is provided in \hyperlink{box:tool_editor}{\texttt{PROMPT}}. It is prompted with examples of various broken tool-call cases and outputs either a corrected tool call, “UNCHANGED” if the original call is valid, or “NO\_VALID\_TOOL\_CALLS” when the output contains only textual explanations.

\NewDocumentCommand{\AlfWorldExample}{}{%
\parbox[c]{\linewidth}{\tiny
\raggedright
\textbf{Task:} \emph{Look at the bowl under the desklamp.}\\[0.15em]
\begin{tabular}{@{}>{\raggedright\arraybackslash}p{0.2em}%
                >{\raggedright\arraybackslash}p{\dimexpr\linewidth-2.0em\relax}@{}}
\TAOrow{T}{I should check the desks to see if there's a desklamp and a bowl nearby. I'll start with desk~1.}
\TAOrow{A}{go to desk~1}
\TAOrow{O}{On the desk~1, you see a desklamp, a keychain, a mug, pens, and a pencil.}
\TAOrow{T}{I found a desklamp on desk~1. I should check if there's a bowl nearby.}
\TAOrow{A}{examine desk~1}
\TAOrow{O}{No bowl found on desk~1.}
\TAOrow{T}{I didn't find a bowl on desk~1. I'll check desk~2 next.}
\TAOrow{A}{go to desk~2}
\TAOrow{O}{On the desk~2, you see an alarm clock, a bowl, and CDs.}
\TAOrow{T}{I found a bowl on desk~2 and can use the desklamp from desk~1.}
\TAOrow{A}{take bowl from desk~2}
\TAOrow{O}{You pick up the bowl from desk~2.}
\TAOrow{T}{I need to return to desk~1 where the desklamp is.}
\TAOrow{A}{go to desk~1}
\TAOrow{O}{The desklamp is available on desk~1.}
\TAOrow{T}{I should use the desklamp to look at the bowl.}
\TAOrow{A}{use desklamp}
\TAOrow{O}{You turn on the desklamp.}
\end{tabular}
}%
}

\NewDocumentCommand{\LifeSpanExample}{}{%
\parbox[c]{\linewidth}{\tiny
\raggedright 
\textbf{Task:} \emph{Find the animal with the longest life span.}\\[0.15em]
\begin{tabular}{@{}>{\raggedright\arraybackslash}p{0.2em}%
                >{\raggedright\arraybackslash}p{\dimexpr\linewidth-2.0em\relax}@{}}
\TAOrow{T}{I need to focus on the animal with the longest lifespan. Since the bee hive is open and contains multiple adult bees, I should focus on the bees first.}
\TAOrow{A}{focus on adult bee}
\TAOrow{O}{Ambiguous request: multiple adult bees are present in the bee hive.}
\TAOrow{T}{I need to determine which animal has the longest lifespan. I will focus on one bee to gather information.}
\TAOrow{A}{focus on adult bee (in bee hive, in greenhouse)}
\TAOrow{O}{Unknown action.}
\TAOrow{T}{I need to examine the bee to determine its lifespan. I will try to look at the bee for more information.}
\TAOrow{A}{look at adult bee (in bee hive, in greenhouse)}
\TAOrow{O}{No known action matches that input.}
\TAOrow{T}{I need to inspect the bee hive more closely to determine information about the bees.}
\TAOrow{A}{look in bee hive}
\TAOrow{O}{Inside the bee hive are several adult bees.}
\TAOrow{T}{The bee hive does not provide lifespan information. I should explore other locations for animals with longer lifespans.}
\TAOrow{A}{go to outside}
\TAOrow{O}{You move to the outside.}
\TAOrow{T}{I need to explore the outside location to find animals with long lifespans.}
\TAOrow{A}{look around}
\TAOrow{O}{You see various animals, including a baby dragonfly, a baby wolf, and a giant tortoise egg.}
\TAOrow{T}{Tortoises are known for long lifespans, so I should examine the giant tortoise egg first.}
\TAOrow{A}{focus on giant tortoise egg}
\TAOrow{O}{You focus on the giant tortoise egg.}
\end{tabular}
}%
}

\NewDocumentCommand{\BabyAIExample}{}{%
\parbox[c]{\linewidth}{\tiny
\raggedright
\textbf{Task:} \emph{go to the red ball.}\\[0.15em]
\begin{tabular}{@{}>{\raggedright\arraybackslash}p{0.2em}%
                >{\raggedright\arraybackslash}p{\dimexpr\linewidth-2.0em\relax}@{}}
\TAOrow{T}{I need to find the red ball, so I'll start by exploring forward.}
\TAOrow{A}{move forward}
\TAOrow{O}{There is a barrier in front of you, you can't move forward.}
\TAOrow{T}{There's a wall ahead. I’ll turn to look for a clearer path toward the red ball.}
\TAOrow{A}{turn right}
\TAOrow{O}{In front of you in this room, you can see several objects: The room has walls around you. You are facing a wall 2 steps away. You are not carrying anything.}
\TAOrow{T}{There's more space in this direction. I’ll move forward to see if I can spot the red ball.}
\TAOrow{A}{move forward}
\TAOrow{O}{In front of you in this room, you can see several objects: The room has walls around you. You are facing a wall 1 steps away. You are not carrying anything.}
\TAOrow{T}{This wall is close. I'll try turning again to scan for the red ball.}
\TAOrow{A}{turn left}
\TAOrow{O}{In front of you in this room, you can see several objects: The room has walls around you. You are facing a wall 1 steps away. You are not carrying anything.}
\TAOrow{T}{I need to find a different direction to explore since turning left and right is not allowed. I'll try moving forward again to see if there's a new path.}
\TAOrow{A}{move forward}
\TAOrow{O}{There is a barrier in front of you, you can't move forward.}
\TAOrow{T}{I need to find a different direction to explore since moving forward is blocked. I'll try turning left to see if there's a new path.}
\TAOrow{A}{turn left}
\TAOrow{O}{In front of you in this room, you can see several objects: There is a grey box 1 3 steps in front of you and 1 steps to your left. There is a grey ball 1 2 steps in front of you and 2 steps to your left. There is a red ball 1 right in front of you 5 steps away. There is a grey key 1 4 steps in front of you and 1 steps to your left. The room has walls around you. You are facing a wall 6 steps away. You are not carrying anything.}
\TAOrow{T}{I can see the red ball now! Let’s head toward it.}
\TAOrow{A}{go to red ball 1}
\TAOrow{O}{In front of you in this room, you can see several objects: There is a red ball 1 right in front of you 1 steps away. There is a grey key 1 0 steps in front of you and 1 steps to your left. The room has walls around you. You are facing a wall 2 steps away. You are not carrying anything. The task is completed.}
\end{tabular}
}%
}

\FloatBarrier
\begin{table*}[t]
\centering
\small
\setlength{\tabcolsep}{6pt}
\renewcommand{\arraystretch}{1.15}
\begin{tabularx}{\linewidth}{c c X}
\toprule
\textbf{Test Set} & \textbf{\#Tasks} & 
\multicolumn{1}{c}{\textbf{Example Task \& Trajectory}} \\
\midrule
\textbf{AlfWorld} & 134 & \AlfWorldExample\\\midrule
\textbf{ScienceWorld} & 90 & \LifeSpanExample \\\midrule
\textbf{BabyAI} & 112 & \BabyAIExample \\
\bottomrule
\end{tabularx}
\caption{\textbf{Overview of embodied agent tasks in the AgentBoard \citep{chang2024agentboard} benchmark}. In the example trajectories, "T", "A", and "O" denote \textbf{T}hought, \textbf{A}ction, and \textbf{O}bservation, respectively.}
\label{tab:example_agentboard}
\end{table*}
\FloatBarrier

\newlength{\BFCLLabelWidth}
\newlength{\BFCLContentShrink}

\setlength{\BFCLLabelWidth}{0.4em}     
\setlength{\BFCLContentShrink}{1.6em}  

\NewDocumentCommand{\bfclsimple}{}{%
\parbox[c]{\linewidth}{\tiny
\raggedright
\strut\vspace*{0.1ex}
\begin{tabular}{@{}>{\raggedright\arraybackslash}p{\BFCLLabelWidth}%
                >{\raggedright\arraybackslash}p{\dimexpr\linewidth-\BFCLContentShrink\relax}@{}}
\TAOrow{U}{Convert 150 Euros to Canadian dollars.}
\TAOrow{T}{currency\_conversion.convert}
\TAOrow{TC}{[currency\_conversion.convert(amount=150, from\_currency='EUR', to\_currency='CAD')]}
\end{tabular}
\strut\vspace*{0.1ex}
}%
}

\NewDocumentCommand{\bfcljava}{}{%
\parbox[c]{\linewidth}{\tiny
\raggedright
\strut\vspace*{0.1ex}
\begin{tabular}{{%
@{}>{\raggedright\arraybackslash}p{\BFCLLabelWidth}%
  >{\raggedright\arraybackslash}p{\dimexpr\linewidth-\BFCLContentShrink\relax}@{}%
}}
\TAOrow{U}{Can I determine if the symbol 'getVersion' is readable in the native function interface library associated with the current object?}
\TAOrow{T}{NFILibrary.isMemberReadable}
\TAOrow{TC}{[NFILibrary.isMemberReadable(symbol='getVersion')]}
\end{tabular}
\strut\vspace*{0.1ex}
}%
}

\NewDocumentCommand{\bfcljavascript}{}{%
\parbox[c]{\linewidth}{\tiny
\raggedright
\strut\vspace*{0.1ex}
\begin{tabular}{{%
@{}>{\raggedright\arraybackslash}p{\BFCLLabelWidth}%
  >{\raggedright\arraybackslash}p{\dimexpr\linewidth-\BFCLContentShrink\relax}@{}%
}}
\TAOrow{U}{Help me reset a state property called 'userSession' to 'null' in a React component?}
\TAOrow{T}{resetStateProperty}
\TAOrow{TC}{[resetStateProperty(stateProperty=\"userSession\")]}
\end{tabular}
\strut\vspace*{0.1ex}
}%
}

\NewDocumentCommand{\bfclmultiple}{}{%
\parbox[c]{\linewidth}{\tiny
\raggedright
\strut\vspace*{0.1ex}
\begin{tabular}{{%
@{}>{\raggedright\arraybackslash}p{\BFCLLabelWidth}%
  >{\raggedright\arraybackslash}p{\dimexpr\linewidth-\BFCLContentShrink\relax}@{}%
}}
\TAOrow{U}{What's the area of a circle with a radius of 10?}
\TAOrow{T}{geometry.area\_circle, plot\_sine\_wave}
\TAOrow{TC}{[geometry.area\_circle(radius=10)]}
\end{tabular}
\strut\vspace*{0.1ex}
}%
}

\NewDocumentCommand{\bfclparallel}{}{%
\parbox[c]{\linewidth}{\tiny
\raggedright
\strut\vspace*{0.1ex}
\begin{tabular}{{%
@{}>{\raggedright\arraybackslash}p{\BFCLLabelWidth}%
  >{\raggedright\arraybackslash}p{\dimexpr\linewidth-\BFCLContentShrink\relax}@{}%
}}
\TAOrow{U}{How to save game progress at stage 7 in easy mode and stage 3 in hard mode?}
\TAOrow{T}{game.save\_progress}
\TAOrow{TC}{[game.save\_progress(stage=7, mode='easy'), game.save\_progress(stage=3, mode='hard')]}
\end{tabular}
\strut\vspace*{0.1ex}
}%
}

\NewDocumentCommand{\bfclparallelmultiple}{}{%
\parbox[c]{\linewidth}{\tiny
\raggedright
\strut\vspace*{0.1ex}
\begin{tabular}{{%
@{}>{\raggedright\arraybackslash}p{\BFCLLabelWidth}%
  >{\raggedright\arraybackslash}p{\dimexpr\linewidth-\BFCLContentShrink\relax}@{}%
}}
\TAOrow{U}{Invest \$2000 in Google and withdraw \$1000 from Apple.}
\TAOrow{T}{investment.withdraw, investment.invest}
\TAOrow{TC}{[investment.invest(company="Google", amount=2000), investment.withdraw(company="Apple", amount=1000)]}
\end{tabular}
\strut\vspace*{0.1ex}
}%
}

\NewDocumentCommand{\bfcllivesimple}{}{%
\parbox[c]{\linewidth}{\tiny
\raggedright
\begin{tabular}{{%
@{}>{\raggedright\arraybackslash}p{\BFCLLabelWidth}%
  >{\raggedright\arraybackslash}p{\dimexpr\linewidth-\BFCLContentShrink\relax}@{}%
}}
\TAOrow{U}{Order me pizza.}
\TAOrow{T}{ChaFod}
\TAOrow{TC}{[ChaFod(TheFod="PIZZA")]}
\end{tabular}
}%
}

\NewDocumentCommand{\bfcllivemultiple}{}{%
\parbox[c]{\linewidth}{\tiny
\raggedright
\begin{tabular}{{%
@{}>{\raggedright\arraybackslash}p{\BFCLLabelWidth}%
  >{\raggedright\arraybackslash}p{\dimexpr\linewidth-\BFCLContentShrink\relax}@{}%
}}
\TAOrow{U}{Can you find me a Family Counselor in Gilroy?}
\TAOrow{T}{Services\_4\_BookAppointment, Services\_4\_FindProvider, Weather\_1\_GetWeather}
\TAOrow{TC}{[Services\_4\_FindProvider(city="Gilroy, CA", type="Family Counselor")]}
\end{tabular}
}%
}

\NewDocumentCommand{\bfclliveparallel}{}{%
\parbox[c]{\linewidth}{\tiny
\raggedright
\begin{tabular}{{%
@{}>{\raggedright\arraybackslash}p{\BFCLLabelWidth}%
  >{\raggedright\arraybackslash}p{\dimexpr\linewidth-\BFCLContentShrink\relax}@{}%
}}
\TAOrow{U}{What's the snow like in the two cities of Paris and Bordeaux?}
\TAOrow{T}{get\_snow\_report}
\TAOrow{TC}{[get\_snow\_report(location="Paris, France"), get\_snow\_report(location="Bordeaux, France")]}
\end{tabular}
}%
}

\NewDocumentCommand{\bfclliveparallelmultiple}{}{%
\parbox[c]{\linewidth}{\tiny
\raggedright
\begin{tabular}{{%
@{}>{\raggedright\arraybackslash}p{\BFCLLabelWidth}%
  >{\raggedright\arraybackslash}p{\dimexpr\linewidth-\BFCLContentShrink\relax}@{}%
}}
\TAOrow{U}{interviewers list for Python and Java}
\TAOrow{T}{get\_interviewer\_list, review\_of\_interviewer}
\TAOrow{TC}{[get\_interviewer\_list(skill="Python"), get\_interviewer\_list(skill="Java")]}
\end{tabular}
}%
}

\NewDocumentCommand{\bfclmultiturnbase}{}{%
\parbox[c]{\linewidth}{\tiny
\raggedright
\begin{tabular}{{%
@{}>{\raggedright\arraybackslash}p{\BFCLLabelWidth}%
  >{\raggedright\arraybackslash}p{\dimexpr\linewidth-\BFCLContentShrink\relax}@{}%
}}
\TAOrow{U}{Hey, I've just filled my car up with 13.2 gallons of fuel. How much is that in liters?}
\TAOrow{TC}{[gallon\_to\_liter(gallon=13.2)]}
\TAOrow{U}{Once you've converted it to liters, please guide me in filling the tank to the max limit. After that's sorted, I need to ensure that all the doors are securely locked and the parking brake is firmly set.}
\TAOrow{TC}{[fillFuelTank(fuelAmount=36.8), lockDoors(unlock=False, door=["driver", "passenger", "rear\_left", "rear\_right"]), activateParkingBrake(mode="engage")]}...
\end{tabular}
}%
}

\NewDocumentCommand{\bfclmultimissfunc}{}{%
\parbox[c]{\linewidth}{\tiny
\raggedright
\strut\vspace*{0.1ex}
\begin{tabular}{{%
@{}>{\raggedright\arraybackslash}p{\BFCLLabelWidth}%
  >{\raggedright\arraybackslash}p{\dimexpr\linewidth-\BFCLContentShrink\relax}@{}%
}}
\TAOrow{U}{I've been thinking of visiting Autumnville for a while now, but I'm not sure how far it is from here in Crescent Hollow. Can you help me figure this out so I can plan my trip accordingly?}
\TAOrow{TC}{[get\_zipcode\_based\_on\_city(city="Crescent Hollow"), get\_zipcode\_based\_on\_city(city="Autumnville"), estimate\_drive\_feasibility\_by\_mileage(distance=100)], [estimate\_distance(cityA="69238", cityB="51479")]}
\TAOrow{U}{Oh, and by the way, there's something else I need help with. I want to calculate the logarithm of the distance you've just told me about, considering a base 10 with a precision of 5 digits. Could you provide me with this value as well?}
\TAOrow{TC}{[logarithm(value=630.0, base=10, precision=5)]]}
\end{tabular}
\strut\vspace*{0.1ex}
}%
}

\NewDocumentCommand{\bfclmultimissparam}{}{%
\parbox[c]{\linewidth}{\tiny
\raggedright
\strut\vspace*{0.1ex}
\begin{tabular}{{%
@{}>{\raggedright\arraybackslash}p{\BFCLLabelWidth}%
  >{\raggedright\arraybackslash}p{\dimexpr\linewidth-\BFCLContentShrink\relax}@{}%
}}
\TAOrow{U}{Hey there, I noticed that all of my car doors seem to have locked themselves up, and with my schedule being pretty tight today, I'm in quite a pinch. I could really use your help to get some of them unlocked.}
\TAOrow{TC}{[lockDoors(unlock=True, door=["driver", "passenger", "rear\_left", "rear\_right"])], [lockDoors(unlock=False, door=["driver", "passenger", "rear\_left", "rear\_right"])]}
\TAOrow{U}{I mean could you help me get all those doors unlocked? It'd be fantastic if you could also switch on the headlights. It's getting a bit darker out here than expected, and visibility's not great!}
\TAOrow{TC}{[lockDoors(unlock=True, door=["driver", "passenger", "rear\_left", "rear\_right"]), setHeadlights(mode="on")]}
\end{tabular}
\strut\vspace*{0.1ex}
}%
}

\NewDocumentCommand{\bfclmultiturnlongcontext}{}{%
\parbox[c]{\linewidth}{\tiny
\raggedright
\strut\vspace*{0.1ex}
\begin{tabular}{{%
@{}>{\raggedright\arraybackslash}p{\BFCLLabelWidth}%
  >{\raggedright\arraybackslash}p{\dimexpr\linewidth-\BFCLContentShrink\relax}@{}%
}}
\TAOrow{U}{It'd be great if you could pop Zeta Corp's stock onto my watchlist. I've come across some fascinating insights into their recent performance that I want to monitor.}
\TAOrow{TC}{[add\_to\_watchlist(stock="ZETA")]}
\TAOrow{U}{With Zeta Corp's stock now on my radar, let's pull up the complete list of stocks I'm watching. I want to double-check that all my chosen stocks are properly listed for my review.}
\TAOrow{TC}{[get\_watchlist()]}
\end{tabular}
\strut\vspace*{0.1ex}
}%
}

\NewDocumentCommand{\bfclliverel}{}{%
\parbox[c]{\linewidth}{\tiny
\raggedright
\begin{tabular}{{%
@{}>{\raggedright\arraybackslash}p{\BFCLLabelWidth}%
  >{\raggedright\arraybackslash}p{\dimexpr\linewidth-\BFCLContentShrink\relax}@{}%
}}
\TAOrow{U}{Hi, could you get me a house to stay for 4 in London?}
\TAOrow{T}{Hotels\_2\_BookHouse, Hotels\_2\_SearchHouse}
\TAOrow{TC}{[Hotels\_2\_SearchHouse(where\_to="London, UK", number\_of\_adults=4)]}
\end{tabular}
}%
}

\NewDocumentCommand{\bfclirrel}{}{%
\parbox[c]{\linewidth}{\tiny
\raggedright
\strut\vspace*{0.1ex}
\begin{tabular}{{%
@{}>{\raggedright\arraybackslash}p{\BFCLLabelWidth}%
  >{\raggedright\arraybackslash}p{\dimexpr\linewidth-\BFCLContentShrink\relax}@{}%
}}
\TAOrow{U}{What defines scientist?}
\TAOrow{T}{get\_historical\_figure\_info}
\end{tabular}
\strut\vspace*{0.1ex}
}%
}

\NewDocumentCommand{\bfclliveirrel}{}{%
\parbox[c]{\linewidth}{\tiny
\raggedright
\strut\vspace*{0.1ex}
\begin{tabular}{{%
@{}>{\raggedright\arraybackslash}p{\BFCLLabelWidth}%
  >{\raggedright\arraybackslash}p{\dimexpr\linewidth-\BFCLContentShrink\relax}@{}%
}}
\TAOrow{U}{dsfsdf}
\TAOrow{T}{calculate\_tax}
\end{tabular}
\strut\vspace*{0.1ex}
}%
}

\FloatBarrier
\begin{table*}[t]
\centering
\small
\setlength{\tabcolsep}{4pt}
\renewcommand{\arraystretch}{0.95}
\begin{tabularx}{\linewidth}{c c c X}
\toprule
\makecell{\textbf{Categories I} \\ (This Work)} 
& \makecell{\textbf{Categories II} \\ (Original Testset)} 
& \makecell{\textbf{\#Used}\\(\#Original)} 
& \multicolumn{1}{c}{\textbf{Example \& Tool-Call}} \\
\midrule

\multicolumn{4}{c}{\textbf{\textit{Non-Live (300)}}} \\
\hdashline[3pt/3pt]
\noalign{\vskip 0.5ex}

\multirow{6}{*}[-12ex]{\textbf{Non-Live}}
& simple              & 50 (400) & \bfclsimple \\
\cmidrule(lr){2-4}
& java                & 50 (100) & \bfcljava \\
\cmidrule(lr){2-4}
& javascript          & 50 (50)  & \bfcljavascript \\
\cmidrule(lr){2-4}
& multiple            & 50 (200) & \bfclmultiple \\
\cmidrule(lr){2-4}
& parallel            & 50 (200) & \bfclparallel \\
\cmidrule(lr){2-4}
& parallel\_multiple  & 50 (200) & \bfclparallelmultiple \\
\midrule

\multicolumn{4}{c}{\textbf{\textit{Single-Turn Live (140)}}} \\
\hdashline[3pt/3pt]
\noalign{\vskip 0.5ex}

\textbf{S.} & live\_simple              & 50 (258)  & \bfcllivesimple \\
\cmidrule(lr){1-4}
\textbf{M.} & live\_multiple            & 50 (1053) & \bfcllivemultiple \\
\cmidrule(lr){1-4}
\textbf{P.} & live\_parallel            & 16 (16)   & \bfclliveparallel \\
\cmidrule(lr){1-4}
\textbf{PM.} & live\_parallel\_multiple  & 24 (24)   & \bfclliveparallelmultiple \\
\midrule

\multicolumn{4}{c}{\textbf{\textit{Multi-Turn (200)}}} \\
\hdashline[3pt/3pt]
\noalign{\vskip 0.5ex}

\textbf{Standard} & multi\_turn\_base          & 50 (200) & \bfclmultiturnbase \\
\cmidrule(lr){1-4}
\multirow{3}{*}[-10ex]{\textbf{Challenge}} & multi\_turn\_miss\_func    & 50 (200) & \bfclmultimissfunc \\
\cmidrule(lr){2-4}
& multi\_turn\_miss\_param   & 50 (200) & \bfclmultimissparam \\
\cmidrule(lr){2-4}
& multi\_turn\_long\_context & 50 (200) & \bfclmultiturnlongcontext \\
\midrule

\multicolumn{4}{c}{\textbf{\textit{Hallucination (118)}}} \\
\hdashline[3pt/3pt]
\noalign{\vskip 0.5ex}

\textbf{Rel.} & live\_relevance     & 18 (18)  & \bfclliverel \\
\cmidrule(lr){1-4}
\multirow{2}{*}[-1.2ex]{\textbf{Irrel.}} & irrelevance         & 50 (240) & \bfclirrel \\
\cmidrule(lr){2-4}
& live\_irrelevance   & 50 (882) & \bfclliveirrel \\

\bottomrule
\end{tabularx}
\caption{\textbf{Overview of tool-calling tasks in the BFCL benchmark} \citep{patil2025bfcl}. 
"U", "T", and "TC" denote the \textbf{U}ser message, available \textbf{T}ools, and the generated \textbf{T}ool \textbf{C}all, respectively.}
\label{tab:example_bfcl}
\end{table*}
\FloatBarrier


\hypertarget{box:embodied_react}{}
\begin{tcolorbox}[
colback=white!10!white,
colframe=EmbodiedColor,
title=Embodied Agent: ReAct-Style,
breakable]

\textbf{SYSTEM}: 

You are a helpful assistant.

\textbf{USER}: 

Your task is to interact with a virtual household simulator to accomplish a specific task. With each interaction, you will receive an observation. Your role is to ... \hl{\{task instruction\}}
\\\\
Here is the example:
\hl{\{example\}}
\\\\
Now, it's your turn. You should perform thoughts and actions to accomplish the goal. Your response should use the following format:
\\\\
Thought: <your thoughts>

Action: <your next action>
\\\\
Your task is: \hl{\{task goal\}}

You are in the middle of a room. Looking quickly around you, ... \hl{\{init observation\}}
\\\\
\hl{\{interaction history\}}
\\\\
The next action could be chosen from these valid actions: \hl{\{valid actions\}}

\end{tcolorbox}


\hypertarget{box:memory_exemplars}{}
\begin{tcolorbox}[
colback=white!10!white,
colframe=EmbodiedColor,
title=DiffuAgent: 3 Memory Exemplars,
breakable]

\textbf{Example 1}:

Memory: (empty)
\\\\
Thought: I should check nearby storage spaces for a spraybottle.

Action: go to cabinet 1

Observation: On the cabinet 1, you see a cloth 1, a soapbar 1, a soapbottle 1.

\medskip
\textbf{Example 2}:

Memory: The agent has checked cabinet~1 and found no spraybottle, then moved to cabinet~2 and discovered it was closed. Cabinet~2 is still unopened and uninspected inside.
\\\\
Thought: I need to open this cabinet to see what's inside.
\\
Action: open cabinet~2
\\
Observation: You open the cabinet~2. The cabinet~2 is open. In it, you see a candle~1, and a spraybottle~2.

\medskip

\textbf{Example 3}:

Memory: The agent has searched cabinet~1 (without finding a spraybottle), opened cabinet~2 (where the spraybottle was found), picked up spraybottle~2, and moved to the toilet~1 (where a soapbottle was present).
\\\\
Thought: It's time to place the spraybottle on the toilet to complete the task.
\\
Action: put spraybottle~2 in/on toilet~1
\\
Observation: You put the spraybottle~2 in/on the toilet~1.
\end{tcolorbox}


\hypertarget{box:memory_update}{}
\begin{tcolorbox}[
colback=white!10!white,
colframe=EmbodiedColor,
title=DiffuAgent: Pre-hoc Memory Update,
breakable]

\textbf{SYSTEM}: 

You are a memory updater.  
Update the memory\_str to reflect what the agent has done and learned so far.  
Include important actions taken, locations visited, and key observations.  
Keep the summary concise, chronological, and consistent.  
Do not invent new facts or omit relevant past actions.  
Write the memory in third-person, concise past tense, like a mission log.

\medskip

\textbf{USER}: 

Memory\_str: \hl{\{previous memory\_str\}}  

Recent\_steps: \hl{\{recent interaction steps\}}

\medskip

Please output the updated Memory\_str only — a short narrative summary of what has been done and observed so far.  
No explanations or formatting other than plain text.

\medskip

Memory\_str: 

\end{tcolorbox}


\hypertarget{box:early_exit}{}
\begin{tcolorbox}[
colback=white!10!white,
colframe=EmbodiedColor,
title=DiffuAgent: Post-hoc Exit Verification,
breakable]

\textbf{SYSTEM}: 

You are a helpful assistant.

\textbf{USER}: 

You will be given a historical scenario in which you are placed in a specific environment with a designated objective to accomplish.  

\#\#\# Task Description: 
Your task is to interact with a virtual household simulator to accomplish a specific task. With each interaction, you will receive an observation. Your role is to ... \hl{\{task instruction\}}
  
\#\#\# Your Objective: 

\hl{\{task goal\}} 

Your Current History:

\hl{\{interaction history\}}

Instructions:

\hl{\{extrinsic early-exit instruction\}}

Do not include any additional text or explanations in your response.

\end{tcolorbox}


\hypertarget{box:tool_call}{}
\begin{tcolorbox}[
colback=white!10!white,
colframe=ToolcallColor,
title=Tool-Call Agent: BFCL default Prompts,
breakable]

\textbf{SYSTEM}:

You are an expert in composing functions. You are given a question and a set of possible functions. Based on the question, you will need to make one or more function/tool calls to achieve the purpose.

If none of the functions can be used, point it out. If the given question lacks the parameters required by the function, also point it out.
You should only return the function calls in your response.
\\\\
If you decide to invoke any of the function(s), you MUST put it in the format of [func\_name1(params\_name1=params\_value1, params\_name2=params\_value2...), func\_name2(params)]
\\\\
You SHOULD NOT include any other text in the response.
\\\\
At each turn, you should try your best to complete the tasks requested by the user within the current turn. Continue to output functions to call until you have fulfilled the user's request to the best of your ability. Once you have no more functions to call, the system will consider the current turn complete and proceed to the next turn or task.
\\\\
Here is a list of functions in JSON format that you can invoke. 
\hl{\{function descriptions\}}

\medskip

\textbf{USER}:

\hl{\{user message\}}

\end{tcolorbox}


\hypertarget{box:tool_selector}{}
\begin{tcolorbox}[
colback=white!10!white,
colframe=ToolcallColor,
title=DiffuAgent: Pre-hoc Tool Selection,
breakable]

\textbf{SYSTEM}: 

You are a tool selector for a function-calling agent.

\medskip

\textbf{Task}:  

Given a user message ([User Message]), the previous tool call ([Tool Call]) and its results ([Tool Execution Results]), you must select a minimum of 3 distinct functions from the provided list.

\medskip

\textbf{Rules}:
\begin{itemize}
  \item Output at least 3 function names, and no more than 10 functions.
  \item Use ONLY names from the provided function list.
  \item Output ONLY function names. No explanations or extra text.
  \item Prioritize the [USER MESSAGE] above all else; use previous tool calls and results only as supplementary context.
\end{itemize}

\medskip

\textbf{USER}: 

Functions:  

\hl{\{available functions\}}

\medskip

\hl{\{interaction history: user message, tool calls, tool execution results\}}

\medskip

Selected Functions:

\end{tcolorbox}

\hypertarget{box:tool_editor}{}
\begin{tcolorbox}[
colback=white!10!white,
colframe=ToolcallColor,
title=DiffuAgent: Tool-Call Editor,
breakable]

\textbf{SYSTEM}: 

You are a strict tool-call format auditor and repairer.

Your task:  
Repair or validate a broken tool-call and output a final call that strictly follows TOOL\_CALL\_FORMAT.

Rules:
\begin{itemize}
  \item If the tool-call is already valid and correct, output UNCHANGED.
  \item If the tool-call is textual explanations, output NO\_VALID\_TOOL\_CALLS.
  \item If the tool-call contains both explanations and tool-calls, remove the explanations and correct the tool-calls.
  \item If the tool-call does not conform to TOOL\_CALL\_FORMAT, repair any format or schema errors and output the corrected tool-call only; do not invent functions or parameters.
\end{itemize}

\medskip

TOOL\_CALL\_FORMAT:  

[func\_name1(param\_name1=param\_value1, param\_name2=param\_value2, ...),

func\_name2(param\_name3=param\_value3, ...)]

\medskip

Examples:  

BROKEN\_TOOL\_CALL 1:  

[cd(folder="academic\_venture")]  

Output:  
UNCHANGED

\medskip

BROKEN\_TOOL\_CALL 2:  

cd(folder="academic\_venture")  

Output:  
[cd(folder="academic\_venture")]

\medskip

BROKEN\_TOOL\_CALL 3:  

{"cd": {"folder": "academic\_venture"}}  

Output:  
[cd(folder="academic\_venture")]

\medskip

BROKEN\_TOOL\_CALL 4:  

The task is now complete.  

Output:  
NO\_VALID\_TOOL\_CALLS

\medskip

BROKEN\_TOOL\_CALL 5:  

The task is now complete. The final tool-call is {"ls": {}}  

Output:  
[ls()]

\medskip

\textbf{USER}: 

BROKEN\_TOOL\_CALL (to be audited and possibly corrected):

\hl{\{model response\}}

\medskip

Now produce the final output according to the rules above.  
No explanations, markdown, or extra text.

\medskip

Output:

\end{tcolorbox}

\end{document}